\documentclass{article}

\usepackage{iclr2026_conference,times}
\iclrfinalcopy 

\newcommand{\E}{\mathbb{E}}

\newcommand{\N}{\mathcal{N}}
\newcommand{\D}{\mathcal{D}}

\newcommand{\I}{\mathrm{I}}
\newcommand{\KL}{\mathrm{KL}}
\newcommand{\sg}{\operatorname{sg}}
\newcommand{\Var}{\operatorname{Var}}

\newcommand{\argmin}{\operatorname*{arg\,min}}
\newcommand{\argmax}{\operatorname*{arg\,max}}

\newcommand{\ours}{\textsc{SJEPA}\xspace}
\newcommand{\Lpred}{\mathcal{L}_{\mathrm{pred}}}
\newcommand{\Lroll}{\mathcal{L}_{\mathrm{roll}}}
\newcommand{\RIB}{\mathcal{R}_{\mathrm{IB}}}
\newcommand{\Rcorr}{\mathcal{R}_{\mathrm{corr}}}
\newcommand{\Csym}{\Omega}
\newcommand{\Cdyn}{\mathcal{C}_{\mathrm{dyn}}}
\newcommand{\Fsym}{F_{\mathcal{E},\alpha}}
\newcommand{\Hpred}{H_{\mathcal{E},\alpha,\phi}}
\newcommand{\Hpredt}{H_{\mathcal{E},\alpha,\phi,\Delta t}}
\newcommand{\Hpredtt}{H_{\mathcal{E},\alpha,\phi,\Delta t_t}}
\newcommand{\epsinfo}{\varepsilon}
\newcommand{\dd}{\mathrm{d}}

\usepackage{amsmath,amssymb,amsthm,mathtools,bm}
\usepackage{booktabs,multirow,array,tabularx}
\usepackage{graphicx}
\usepackage{xcolor}
\usepackage{microtype}
\usepackage{url}
\usepackage{xspace}
\usepackage[hypertexnames=false]{hyperref}
\usepackage{enumitem}
\usepackage{algorithm}
\usepackage{algorithmic}
\usepackage{float}
\usepackage{tikz}
\usetikzlibrary{arrows.meta,positioning,fit,calc}

\hypersetup{colorlinks=true,citecolor=blue,linkcolor=blue,urlcolor=blue}

\newtheorem{definition}{Definition}[section]
\newtheorem{proposition}[definition]{Proposition}
\newtheorem{theorem}[definition]{Theorem}

\newtheorem{remark}[definition]{Remark}

\title{SJEPA: Learning Elegant Latent Dynamics with Hybrid Symbolic--Neural Predictors}

\author{
Yongchao Huang\thanks{Email: \texttt{yongchao.huang@abdn.ac.uk}}\\\hspace{0.5cm}23/07/2026
}

\date{23 July 2026}

\begin{document}

\maketitle
\lhead{} 

\begin{abstract}
Joint-embedding predictive architectures learn abstract states by predicting target embeddings from context embeddings, but their transition models are typically opaque neural maps. We introduce \ours, a reconstruction-free JEPA framework that learns predictive representations whose induced dynamics admit compact symbolic descriptions. Its hybrid transition combines a symbolic law with a regularised neural correction for dynamics that the selected grammar cannot express adequately. The central principle is to learn the simplest adequate dynamics: representation constraints restrict learning to informative, non-collapsed predictive coordinates, while operator compression favours the lowest-complexity symbolic--neural transition that remains predictively adequate. We formalise this principle through induced-dynamics complexity, analyse the non-identifiability of predictive coordinates, and show that unconstrained operator compression creates a direct shortcut to representation collapse. The framework supports both alternating representation--equation learning and symbolic dynamics fitted to fixed representations. In controlled pendulum experiments, joint representation--equation learning discovers substantially simpler symbolic dynamics with lower long-horizon rollout error and divergence than post-hoc fitting, while an unconstrained one-step diagnostic realises the collapse shortcut predicted by the theory. Under grammar misspecification, correction regularisation preserves the representable symbolic mechanism and encourages the neural component to focus on residual dynamics. The results demonstrate a controllable trade-off among predictive fidelity, representation quality, symbolic parsimony, and symbolic--neural allocation. SJEPA provides a complementary direction within the JEPA family and may be useful in applications where compact latent dynamics, explicit structural constraints, or controlled residual modelling are desirable.
\end{abstract}

\section{Introduction}

Joint-embedding predictive architectures (JEPAs) learn by predicting representations of missing or future observations rather than reconstructing all observation details \citep{lecun2022path,assran2023ijepa,bardes2024vjepa,assran2025vjepa2}. This design separates predictive semantics from pixel-level variability and provides a natural foundation for general-purpose world models. Yet the transition mechanism itself is usually represented by a neural predictor. Such a predictor can be accurate without revealing which variables interact, how actions alter the future, or whether the learned coordinates support a concise dynamical description.

This paper asks a different question from standard representation learning: \emph{can a JEPA learn not only predictive states, but elegant dynamics over those states?} We use ``elegant'' in a precise operational sense: an elegant transition is a compact and parsimonious law that remains adequate for prediction. The objective is therefore not the shortest possible equation. An equation that is too simple underfits; a representation that is made trivial merely to simplify the equation is also unacceptable. The target is the \emph{simplest adequate governing law} for an informative predictive state. The core principle is that operator compression should select predictive coordinates whose induced dynamics are simple yet adequate, while representation constraints prevent that simplicity from being achieved through collapse.

We introduce Symbolic JEPA (SJEPA\footnote{We use the unhyphenated abbreviation \ours because the framework is general-purpose rather than task-specific.}), a JEPA whose latent transition is decomposed into a symbolic governing law and a neural correction. Given a context embedding $Z_C$ and target-side information $\epsinfo$, with $Z_T$ providing the prediction target, the hybrid predictor is
\begin{equation}
\label{eq:intro-hybrid}
\tag{Eq.\ref{eq:hybrid}}
\widehat Z_T
=
f_{\mathcal E,\alpha}(Z_C,\epsinfo)
+
c_\phi(Z_C,\epsinfo).
\end{equation}
Here, $\epsinfo$ identifies the target or transition, for example through a target location, temporal offset, or action. The symbolic structure $\mathcal E$ and coefficients $\alpha$ capture the dominant reusable dynamics, while $c_\phi$ corrects effects that the selected grammar does not represent adequately. This decomposition acts on the transition operator rather than partitioning the latent representation.

SJEPA distinguishes representation compression from operator compression. The former determines which future-relevant state is retained; the latter seeks a concise law governing that state. These objectives must be coupled carefully because an unconstrained pressure for simple dynamics can encourage the encoder to erase information until the transition becomes trivial. We therefore formulate \ours as constrained operator compression:
\begin{equation}
\label{eq:intro-constrained}
\tag{Eq.\ref{eq:constrained}}
\begin{aligned}
\min_{\theta,\bar\theta,\mathcal E,\alpha,\phi}\quad
&
\Csym(\mathcal E)
+
\lambda_{\mathrm c}\Rcorr(\phi)
\\
\text{subject to}\quad
&
\Lpred(\theta,\bar\theta,\mathcal E,\alpha,\phi)
\le
\delta_{\mathrm{pred}},
\\
&
(\theta,\bar\theta)
\in
\Theta_{\mathrm{repr}}.
\end{aligned}
\end{equation}
The objective favours a compact symbolic law while discouraging unnecessary reliance on the correction. The predictive constraint prevents underfitting, whereas $\Theta_{\mathrm{repr}}$ restricts the context and target encoders to informative, predictively adequate, and non-collapsed representations. This formulation induces the dynamics-complexity functional
$\Cdyn(E_\theta,E_{\bar\theta})$: the minimum symbolic-plus-correction complexity required for adequate prediction in the coordinates selected by the encoders.

Our contributions are:
\begin{enumerate}[leftmargin=*,itemsep=2pt]
\item \textit{Symbolic and hybrid JEPA dynamics.} We replace the opaque neural predictor in JEPA with a compact symbolic law, together with an optional neural correction for dynamics that cannot be expressed adequately by the selected symbolic grammar.

\item \textit{Learning the simplest adequate dynamics.}
We introduce operator compression as a complement to representation learning:
the representation must remain informative and non-collapsed, while its
transition should be as compact as possible without sacrificing predictive
adequacy.

\item \textit{A principled and modular learning framework.} We develop objectives that control symbolic complexity and discourage the neural correction from unnecessarily absorbing representable dynamics. The framework supports both alternating representation--equation learning and symbolic dynamics fitted to frozen pretrained encoders.

\item \textit{Theoretical and empirical validation.} We explain why predictive coordinates are not unique, show how dynamics compression can encourage representation collapse, and validate these effects experimentally. Joint learning discovers substantially simpler latent dynamics with more accurate and less divergent symbolic rollouts than post-hoc equation fitting, while correction regularisation preserves the dominant symbolic mechanism under grammar misspecification.

\item \textit{Extensions to uncertainty and control.} We provide Bayesian and action-conditioned formulations that support uncertainty-aware latent dynamics, sensitivity analysis, local linearisation, and model-based planning.
\end{enumerate}

\section{Related Works}

\paragraph{Joint-embedding predictive architectures.}
The JEPA family has expanded across modalities and learning objectives. I-JEPA introduced masked latent prediction for images \citep{assran2023ijepa}, while MC-JEPA jointly learns motion and content features and V-JEPA extends feature prediction to video \citep{bardes2023mcjepa,bardes2024vjepa}. Audio-JEPA, Point-JEPA, and 3D-JEPA adapt the framework to audio and three-dimensional data, and VL-JEPA extends latent prediction to vision--language learning \citep{tuncay2025audiojepa,saito2024pointjepa,hu20243djepa,chen2025vljepa}. Other variants move toward structured world modelling and decision making: ACT-JEPA jointly predicts actions and latent observation sequences, V-JEPA~2 develops action-conditioned video world models for planning, and C-JEPA introduces object-level latent interventions for causal reasoning \citep{vujinovic2025actjepa,assran2025vjepa2,nam2026causaljepa}. LeJEPA provides a leaner and theoretically grounded training objective based on isotropic representation regularisation \citep{balestriero2025lejepa}. Complementing these deterministic formulations, Huang's variational JEPA (VJEPA) develops a probabilistic JEPA framework that learns predictive distributions over future latent states and connects JEPA with predictive-state representations and Bayesian filtering \citep{huang2026vjepa}. Despite this diversity, most variants retain flexible neural predictors or primarily modify the representation objective, modality, or conditioning structure. \ours is complementary: it asks whether the induced latent transition itself can be compressed into a compact symbolic law, with a controlled neural correction for residual dynamics.

\paragraph{Symbolic regression and governing-equation discovery.}
Symbolic regression searches over analytical expressions rather than assuming a fixed parametric form \citep{koza1992genetic,schmidt2009distilling}. SINDy identifies sparse governing dynamics within a user-specified library of candidate functions, while SINDYc extends this formulation to systems with external inputs and control \citep{brunton2016sindy,brunton2016sindyc}. More recent methods distil symbolic relations from trained neural models, use neural generators or pretrained transformers to propose expressions, or employ scalable evolutionary search \citep{cranmer2020symbolic,petersen2021deep,biggio2021neural,cranmer2023pysr}. Most such methods operate on a supplied set of explanatory variables; in governing-equation discovery, the state coordinates and either their derivatives or transition data are typically observed or selected beforehand. \ours instead couples symbolic equation discovery with learning the predictive coordinates in which those equations are expressed.

\paragraph{Joint coordinate and equation discovery.}
Several important precedents learn coordinates together with structured dynamics. SINDy autoencoders jointly learn a nonlinear coordinate transformation, a reconstruction map, and sparse latent governing equations \citep{champion2019coordinates}. T-SHRED combines transformer-based temporal encoding with a SINDy-attention mechanism that regularises and interprets latent dynamics when forecasting full systems from sparse sensor histories \citep{yermakov2025tshred}. DYSCO uses multiple noisy views and temporal contrastive learning to jointly recover latent trajectories and structured dynamics, with identification guarantees up to an affine indeterminacy and symbolic recovery within that gauge \citep{muratore2026dysco}. These works establish that representation learning and equation discovery can be coupled. \ours addresses a complementary setting through a reconstruction-free JEPA context--target interface, optional target-side information, and an explicit symbolic law plus neural correction. It further uses minimum adequate operator complexity to guide representation learning while imposing representation constraints to prevent trivial dynamics from being obtained through collapse.

\paragraph{Learned coordinates for dynamics.}
Latent-dynamics models combine compressed representations with learned transition models for prediction and simulation. World Models, for example, learns compact visual states together with recurrent latent dynamics, while Neural ODEs parameterise continuous-time hidden-state evolution and can be used within latent-variable models \citep{ha2018worldmodels,chen2018neuralode}. Koopman-based methods pursue a more specific form of simplification by seeking observables or coordinates in which nonlinear dynamics evolve linearly, or approximately linearly, within a suitable invariant subspace \citep{brunton2016koopman,lusch2018koopman}. 
\ours adopts a different but compatible inductive bias: the latent transition need not be linear, but should admit a compact symbolic description, with a controlled neural correction where the symbolic grammar is insufficient. Linear latent dynamics remain a possible special case of this broader symbolic operator class.

\paragraph{Information bottlenecks and non-collapse.}
The information bottleneck seeks a compressed representation that preserves information relevant to a target, with the variational information bottleneck providing a practical latent-variable formulation \citep{tishby2000ib,alemi2017vib}. Recent analysis of V-JEPA through the information bottleneck and predictive information bottleneck distinguishes the removal of non-predictive nuisance information from the separate problem of avoiding representation collapse; predictable nuisance information may still be retained \citep{huang2026ibvjepa}. Non-contrastive representation-learning methods address collapse and redundancy through mechanisms such as variance preservation, cross-correlation control, architectural asymmetry, target networks, and distributional regularisation \citep{bardes2022vicreg,zbontar2021barlow,grill2020byol,balestriero2025lejepa}. These methods are not equivalent to one another or to an information-theoretic bottleneck. In \ours, $\RIB$ is therefore an umbrella notation for the representation-side requirement that the learned state remain predictive, informative, and non-collapsed. This requirement is particularly important because operator compression introduces a direct shortcut: the encoder can make the transition trivial by making the representation nearly constant. Our experiments instantiate $\RIB$ with a VICReg-style regulariser and use an unconstrained one-step diagnostic to show that the predicted fixed-point collapse shortcut can be reached in practice.

\paragraph{Mechanistic and neuro-symbolic world models.}
Mechanistic World Models propose organising learned world knowledge around reusable explanatory mechanisms rather than predictive mappings alone \citep{posner2026mechanistic}. Neuro-symbolic JEPA explores a complementary connection between JEPA and symbolic reasoning: rule-informed JEPA uses explicit logical rules to shape the latent energy landscape and uses the learned continuous space to support differentiable rule discovery \citep{huang2026neurosymbolicjepa}. \ours addresses a different form of symbolic structure. It does not primarily inject logical rules into the representation or discover rules relating symbolic concepts; instead, it learns a compact symbolic operator that describes how a predictive latent state evolves from context to target, optionally conditioned on time, actions, or other side information. A regularised neural correction accounts for residual dynamics that the selected grammar cannot express adequately. The resulting equations should be interpreted as compact predictive mechanisms rather than automatically as physical laws. Predictive, representation, complexity, and correction constraints are required to prevent an apparently simple symbolic description from being obtained through underfitting, collapse, or delegation of the dynamics to an unrestricted neural component.

\section{\ours: Symbolic and Hybrid Latent Dynamics}

\subsection{Problem formulation and notation}

A training example is $(X_C,X_T,\epsinfo)\sim\D$, where $X_C$ is a context observation or history, $X_T$ is a target observation or future segment, and $\epsinfo$ denotes (possibly empty) side information supplied to the predictor. $\epsinfo$ may encode target location, mask geometry, time offset, action, goal, intervention, or other conditioning variables.
The context and target encoders produce
\begin{equation}
\label{eq:encoders}
Z_C
=
E_\theta(X_C),
\qquad
Z_T
=
E_{\bar\theta}(X_T),
\end{equation}
where both embeddings lie in a common latent space\footnote{Although the context and target encoders have different parameter values during EMA training, they are treated as two parameterisations of the same ambient latent coordinate space $\mathcal Z$. The target encoder supplies slowly moving prediction targets rather than defining a separate latent coordinate chart. For temporal prediction, the free-rollout objective additionally trains $\Hpred$ on its own recursively generated outputs.}. The target encoder may be an exponential-moving-average (EMA) copy of the context encoder; in the frozen-encoder mode, both may instead be fixed copies of a pretrained encoder. A standard deterministic JEPA predictor can be written as
\begin{equation}
\widehat Z_T
=
g_\psi(Z_C,\epsinfo),
\end{equation}
where $\psi$ denotes the parameters of an unrestricted neural predictor. In \ours, the predictor is\footnote{Equation~\eqref{eq:hybrid} gives the general, direct-transition realisation of \ours, in which the symbolic and neural components directly predict the target embedding. For continuous-time temporal systems (e.g. a dynamical system), the same decomposition may instead parameterise a latent vector field (e.g. gradient)
\begin{equation*}
\label{eq:hybrid-vector-field}
V_{\mathcal E,\alpha,\phi}(z,\epsinfo)
=
F_{\mathcal E,\alpha}(z,\epsinfo)
+
c_\phi(z,\epsinfo),
\end{equation*}
from which the complete transition is constructed by an integration operator:
\begin{equation*}
\label{eq:integrated-hybrid-transition}
\Hpredt(z,\epsinfo)
=
\mathcal I_{\Delta t}
\left(
z,
V_{\mathcal E,\alpha,\phi},
\epsinfo
\right).
\end{equation*}
Our later experiments use the forward-Euler realisation
$
\Hpredt(z)
=
z
+
\Delta t
\left[
F_{\mathcal E,\alpha}(z)
+
c_\phi(z)
\right].
$
Thus, the symbolic--neural decomposition may apply either to a direct transition map or to a vector field from which the transition map is constructed.}
\begin{equation}
\label{eq:hybrid}
\Hpred(Z_C,\epsinfo)
=
\Fsym(Z_C,\epsinfo)
+
c_\phi(Z_C,\epsinfo),
\end{equation}
where $\Fsym\equiv F_{\mathcal E,\alpha}$ is a symbolic expression with structure $\mathcal E$ and coefficients $\alpha$, and $c_\phi$ is a neural corrector for dynamics not captured adequately by the symbolic grammar. Setting $c_\phi\equiv0$ recovers the purely symbolic model.

The structure $\mathcal E$ is a typed syntax tree drawn from a grammar $\mathfrak E$ of arithmetic, low-order polynomial, transcendental, and domain-specific primitives. Its structural complexity is
\begin{equation}
\label{eq:omega}
\Csym(\mathcal E)
=
\sum_{v\in\operatorname{nodes}(\mathcal E)}
w_{\operatorname{type}(v)},
\end{equation}
where $w_{\operatorname{type}(v)}\ge0$ assigns a complexity cost to each operator or primitive. This weighted score favours concise expressions while allowing more expressive or structurally complex primitives to receive larger penalties.

In the differentiable sparse-library implementation used in our experiments, $\mathcal E$ is represented by the active library support and $\alpha$ by its coefficients. In that case, the same notation $\Csym(\mathcal E)$ denotes the sum of prespecified complexity weights over active output--term pairs; a smooth coefficient-dependent surrogate is used during optimisation and the thresholded support is used for final reporting. Alternatively, the structural score may be replaced by a minimum-description-length criterion\footnote{The minimum description length (MDL) principle selects the model that minimises the total code length required to describe both the model and the data not explained by it. In our setting, this balances the description length of the symbolic structure and its coefficients against the remaining predictive error. MDL therefore does not favour the shortest expression unconditionally; it favours a concise expression that accounts adequately for the observed latent transitions.}.

The preceding definitions specify the representation and hybrid transition model. Section~\ref{sec:elegant} introduces predictive adequacy, symbolic complexity, and correction control, and combines them to formulate elegant-dynamics learning as constrained operator compression.

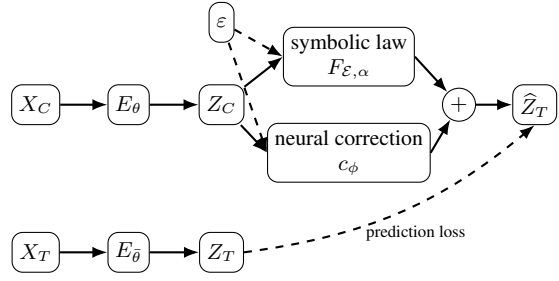
\begin{figure}[t]
\centering
\begin{tikzpicture}[
scale=0.91,
transform shape,
box/.style={
draw,
rounded corners,
minimum height=0.58cm,
align=center,
inner sep=3pt,
font=\small
},
arr/.style={-{Latex[length=2mm]},thick},
sum/.style={draw,circle,inner sep=1.5pt,font=\small}
]
\node[box] (xc) at (0,1.35) {$X_C$};
\node[box] (enc) at (1.35,1.35) {$E_\theta$};
\node[box] (zc) at (2.70,1.35) {$Z_C$};
\node[box] (eps) at (2.70,2.55) {$\epsinfo$};
\node[box] (sym) at (4.55,2.05) {symbolic law\\$\Fsym$};
\node[box] (corr) at (4.55,0.65) {neural correction\\$c_\phi$};
\node[sum] (plus) at (6.15,1.35) {$+$};
\node[box] (pred) at (7.25,1.35) {$\widehat Z_T$};
\node[box] (xt) at (0,-0.80) {$X_T$};
\node[box] (tenc) at (1.35,-0.80) {$E_{\bar\theta}$};
\node[box] (zt) at (2.70,-0.80) {$Z_T$};

\draw[arr] (xc)--(enc);
\draw[arr] (enc)--(zc);
\draw[arr] (zc)--(sym.west);
\draw[arr] (zc)--(corr.west);
\draw[arr,dashed] (eps)--(sym.west);
\draw[arr,dashed] (eps)--(corr.west);
\draw[arr] (sym.east)--(plus);
\draw[arr] (corr.east)--(plus);
\draw[arr] (plus)--(pred);
\draw[arr] (xt)--(tenc);
\draw[arr] (tenc)--(zt);
\draw[arr,dashed] (zt.east) to[bend right=16]
node[
below,
font=\scriptsize,
pos=0.56,
yshift=-5pt
]{prediction loss} (pred.south);
\end{tikzpicture}
\caption{\ours decomposes the latent transition operator into a compact symbolic governing law and a neural correction for dynamics. The decomposition acts on the transition operator rather than partitioning the latent representation.}
\label{fig:architecture}
\end{figure}

\subsection{Why a hybrid predictor?}

A purely symbolic law is directly inspectable but may fail when the chosen grammar omits relevant interactions\footnote{Discovering interactions, especially higher-order ones, is a challenging problem in statistical learning and can be particularly difficult in physical and dynamical systems.} or when the latent coordinates retain unresolved variability. A purely neural predictor is flexible but may absorb predictive regularities without exposing them. The hybrid model therefore seeks a disciplined division of labour: the symbolic component represents the dominant reusable mechanism, while the regularised correction accounts for effects that a compact law does not capture adequately. In a physical analogy, $\Fsym$ may represent a governing relation, while $c_\phi$ may capture friction, unresolved interactions, approximation error, observation effects, or representation mismatch.

This interpretation does not imply that the two components form a unique or intrinsically physical decomposition. Prediction loss alone cannot determine how a transition should be divided between $\Fsym$ and $c_\phi$. The symbolic-complexity and correction penalties instead encode a modelling preference: use a concise symbolic explanation where adequate, and reserve the correction for the remaining predictive structure. Section~\ref{sec:elegant} formalises this allocation and its limitations.

\subsection{The encoder as a learned transform}

In the end-to-end mode, the encoder pair can be interpreted as learning coordinates in which the context--target relation admits a simple symbolic description. In the frozen-encoder mode, the same interpretation instead asks whether the pretrained coordinates already expose such a relation. Let $\mathcal T_{\epsinfo}$ denote the observation-space target mechanism associated with side information $\epsinfo$, written schematically as
\begin{equation}
X_T
\approx
\mathcal T_{\epsinfo}(X_C).
\end{equation}
For temporal tasks, $\mathcal T_{\epsinfo}$ is a dynamical transition; for masked or spatial prediction, it denotes the corresponding target-generating relation. In stochastic settings, it represents the relevant conditional target mechanism rather than a deterministic map.

There are then two routes from the context observation to the target latent representation. The first applies the observation-space target mechanism and subsequently encodes the resulting target; the second encodes the context and predicts the target embedding from the context embedding together with the same side information. \ours seeks consistency between these routes:
\begin{equation}
\label{eq:commute}
E_{\bar\theta} \left(
\mathcal T_{\epsinfo}(X_C)
\right)
\approx
\Hpred \left(
E_\theta(X_C),
\epsinfo
\right).
\end{equation}
Thus, $\epsinfo$ identifies the relevant target relation and is supplied explicitly, together with the context embedding $Z_C=E_\theta(X_C)$, to the latent predictor $\Hpred$.

The analogy is to selecting Laplace, Fourier, canonical, or other transformed coordinates that simplify a governing relation. The claim is not that every system becomes linear or that the resulting coordinates must coincide with physical state variables. Rather, \textit{the representation should expose informative, non-collapsed coordinates in which an adequate symbolic context--target operator has low description complexity}. For temporal tasks, this operator is the latent transition law. Such coordinates may be non-unique and need not remain affinely aligned with the original state, as observed later in Experiment~1.

\section{Learning Elegant Dynamics by Operator Compression}
\label{sec:elegant}

\subsection{Predictive adequacy and operator complexity}

For a pointwise latent-space discrepancy $d$, the one-step predictive risk is
\begin{equation}
\label{eq:lpred}
\Lpred
=
\E_{\D}
\left[
d \left(
\Hpred(Z_C,\epsinfo),
\sg(Z_T)
\right)
\right],
\end{equation}
where $\sg(\cdot)$ stops gradients through the target embedding. The discrepancy $d$ may, for example, be a squared Euclidean or cosine distance. For temporal prediction, one-step accuracy may be supplemented by the multi-step rollout loss $\Lroll$ defined in Equation~\eqref{eq:rollout-loss} of Appendix~\ref{app:objectives}.

The neural correction should improve predictive fidelity without absorbing reusable structure that can be represented adequately by the symbolic law. We therefore regularise both its output magnitude and parameter capacity:
\begin{equation}
\label{eq:rcorr}
\Rcorr(\phi)
=
\E_{\D}
\left[
\left\|
c_\phi(Z_C,\epsinfo)
\right\|_2^2
\right]
+
\tau_\phi\|\phi\|_2^2,
\end{equation}
where $\tau_\phi\ge0$ controls parameter regularisation. The first term discourages large corrections at the prediction level, while the second limits unnecessary complexity in the correction model.

\begin{definition}[Operator-complexity score]
For a fixed latent normalisation, symbolic grammar, and correction class, define the operator-complexity score of the hybrid transition
$\Hpred=\Fsym+c_\phi$ in Equation~\eqref{eq:hybrid} by
\begin{equation}
\label{eq:gamma}
\Gamma(\mathcal E,\phi)
=
\Csym(\mathcal E)
+
\lambda_{\mathrm c}\Rcorr(\phi),
\end{equation}
where $\lambda_{\mathrm c}\ge0$ controls the trade-off between symbolic parsimony and reliance on the neural correction.
\end{definition}

\begin{definition}[Elegant dynamics]
Fix an admissible prediction threshold $\delta_{\mathrm{pred}}$ and representation class $\Theta_{\mathrm{repr}}$. An \emph{elegant} \ours solution is a feasible encoder--predictor tuple that minimises $\Gamma$. It therefore realises the simplest adequate transition within the chosen symbolic grammar and correction class, rather than the simplest transition without qualification.
\end{definition}

This definition balances predictive fidelity, symbolic parsimony, and reliance on the correction. Predictive adequacy prevents a short but inaccurate law from being considered elegant. 

\subsection{Representation compression versus operator compression}

Representation learning and dynamics learning play complementary roles, but they answer different questions:
\begin{equation*}
\textit{What future-relevant state should be retained?}
\end{equation*}
and
\begin{equation*}
\textit{What is the simplest adequate law governing that state?}
\end{equation*}

Representation learning must avoid collapse, redundancy, and the removal of variables required for future prediction, while discarding observation-specific nuisance information where possible. The representation module may therefore implement a full information bottleneck \citep{huang2026ibvjepa} or a practical surrogate that preserves an informative, non-collapsed predictive state.

Dynamics learning faces a different trade-off. If the transition is insufficiently compressed, a flexible symbolic expression or neural correction may reproduce the data accurately while remaining opaque. If it is compressed too aggressively, the resulting law may underfit, extrapolate poorly, or pressure the encoder towards artificially simple coordinates. A misspecified grammar may omit mechanisms required for prediction, while an overly expressive correction may absorb the structure that the symbolic component is intended to reveal. The objective is therefore not the simplest possible dynamics, but the simplest law that remains predictively adequate.

Given an admissible representation, the dynamics module controls the induced transition through the symbolic-complexity term $\Csym(\mathcal E)$ and correction penalty $\Rcorr(\phi)$. Thus, \ours does not seek simplicity by discarding the state itself; it seeks an informative representation whose evolution admits a concise and adequate description.

Table~\ref{tab:compression-main} summarises the distinct failure regimes associated with the two forms of compression. The slogan ``compress the dynamics, not the representation'' should therefore be interpreted precisely: the operator-complexity objective must not achieve simplicity by erasing the state\footnote{For example, a collapsed \ours model may map every context and target observation to the same arbitrary vector $z_0$, so that
$E_\theta(X_C)=E_{\bar\theta}(X_T)=z_0$. An identity predictor $\Hpred(z,\epsinfo)=z$, or any transition having $z_0$ as a fixed point, then predicts every target embedding perfectly. Collapse is characterised by the absence of variation or information in the representation, not by whether its constant value is zero.}. Representation compression may discard nuisance information only while preserving predictive sufficiency and non-collapse; operator compression may simplify the transition only while preserving predictive adequacy.

\begin{table}[t]
\centering
\small
\caption{Representation and operator compression address different objectives and failure modes. Operator compression is meaningful only within an admissible predictive representation class.}
\label{tab:compression-main}
\begin{tabularx}{\columnwidth}{p{0.34\columnwidth}X}
\toprule
Regime & Consequence \\
\midrule
Too much representation compression
&
Collapse or loss of future-relevant state information.
\\
Too little representation control
&
Retention of nuisance variation, redundant coordinates, or degenerate predictive shortcuts.
\\
Too little operator compression
&
Accurate but opaque dynamics; the symbolic expression or neural correction absorbs unnecessary complexity.
\\
Too much operator compression
&
Underfit or brittle laws, poor extrapolation, and pressure towards artificially simple coordinates.
\\
Balanced compression
&
Informative predictive states governed by a concise and adequate transition law.
\\
\bottomrule
\end{tabularx}
\end{table}

\subsection{Constrained formulation and induced-dynamics complexity}

Let $\Theta_{\mathrm{repr}}$ denote the admissible set of encoder pairs satisfying the selected predictive-information and anti-collapse requirements. The ideal \ours objective is
\begin{equation}
\label{eq:constrained}
\begin{aligned}
\min_{\theta,\bar\theta,\mathcal E,\alpha,\phi}\quad
&
\Gamma(\mathcal E,\phi)
\\
\mathrm{s.t.}\quad
&
\Lpred
\le
\delta_{\mathrm{pred}},
\\
&
(\theta,\bar\theta)
\in
\Theta_{\mathrm{repr}}.
\end{aligned}
\end{equation}
The predictive constraint prevents operator simplicity from being obtained through underfitting, while the representation constraint prevents it from being obtained through collapsed or uninformative coordinates.

For fixed encoders, define the induced-dynamics complexity
\begin{equation}
\label{eq:cdyn}
\Cdyn(E_\theta,E_{\bar\theta})
=
\inf_{\mathcal E,\alpha,\phi}
\left\{
\Gamma(\mathcal E,\phi)
:
\Lpred
\le
\delta_{\mathrm{pred}}
\right\}.
\end{equation}
This quantity asks how simple the latent transition can be once the representation coordinates have been fixed. It searches over symbolic structures $\mathcal E$, coefficients $\alpha$, and neural corrections $\phi$ that satisfy the predictive requirement, and returns the smallest symbolic-plus-correction complexity among them. A low value indicates that adequate prediction requires a compact symbolic expression and little reliance on the correction. If no predictor satisfies the constraint, then
$
\Cdyn(E_\theta,E_{\bar\theta})
=
+\infty.
$
For temporal applications in which recursive accuracy is part of the required notion of adequacy, the feasible set in Equation~\eqref{eq:cdyn} may additionally impose\footnote{We retain the one-step constraint in the task-general definition because recursive rollout is not defined for every JEPA context--target relation.}
$\Lroll\le\delta_{\mathrm{roll}}$.

\begin{remark}[Relative nature of induced-dynamics complexity]
\label{rem:relative-complexity}
The quantity $\Cdyn$ is not an intrinsic coordinate-invariant property of the underlying physical system. It is defined relative to the admissible representation class, latent normalisation, symbolic grammar and primitive weights, correction function class, and data distribution used to evaluate predictive and correction risks. In particular, arbitrary rescaling of the latent coordinates can change coefficient magnitudes, correction energy, and thresholded symbolic support. The admissible class $\Theta_{\mathrm{repr}}$ should therefore include an explicit scale convention, such as bounded per-coordinate variance, unit-variance normalisation, or whitening. The experiments enforce this convention through the unit-scale variance target in $\RIB^{\mathrm{exp}}$.
\end{remark}

Representation selection may then be written as
\begin{equation}
\label{eq:encoder-selection}
\min_{(\theta,\bar\theta)\in\Theta_{\mathrm{repr}}}
\Cdyn(E_\theta,E_{\bar\theta}).
\end{equation}
The inner optimisation finds the simplest adequate transition for each fixed representation, while the outer optimisation selects, among informative and non-collapsed representations, coordinates whose induced dynamics are easier to describe. Equation~\eqref{eq:encoder-selection} formalises the central principle of \ours: a representation is valuable not only because it supports prediction, but also because its induced dynamics admit a concise and adequate governing law.

\subsection{Predictive non-identifiability and coordinate selection}

\begin{proposition}[Predictive coordinates are non-identifiable]
\label{prop:nonident}
Let $E$ map context and target observations into a common latent space
$\mathcal Z$, and suppose that
\begin{equation}
E(X_T)
=
F(E(X_C),\epsinfo)
\end{equation}
almost surely, where
$F:\mathcal Z\times\mathcal S_{\epsinfo}\rightarrow\mathcal Z$.
For any bijection
$h:\mathcal Z\rightarrow\widetilde{\mathcal Z}$,
define
\begin{equation}
\widetilde E
=
h\circ E
\end{equation}
and
\begin{equation}
\widetilde F(\widetilde z,\epsinfo)
=
h \left(
F(h^{-1}(\widetilde z),\epsinfo)
\right).
\end{equation}
Then
\begin{equation}
\widetilde E(X_T)
=
\widetilde F(\widetilde E(X_C),\epsinfo)
\end{equation}
almost surely. However, the symbolic description complexities of $F$ and $\widetilde F$ need not agree.
\end{proposition}

Intuitively, the proposition says that a predictive latent representation can be rewritten in any invertible coordinate system without changing what the model predicts. The encoder first expresses the state in the new coordinates through $h$, while the transformed predictor undoes this change, applies the original transition, and maps the result back. Prediction alone therefore cannot distinguish between these coordinate systems. Their symbolic descriptions, however, may differ substantially: a transition that is simple in one coordinate system may become complicated in another.

The transformed transition follows the coordinate route
\begin{equation*}
\widetilde z_C
\xrightarrow{\,h^{-1}\,}
z_C
\xrightarrow{\,F(\cdot,\epsinfo)\,}
z_T
\xrightarrow{\,h\,}
\widetilde z_T.
\end{equation*}
It maps the transformed context state back to the original coordinates, applies the original transition, and maps the predicted target state into the transformed coordinates.

The proof is given in Appendix~\ref{app:proof-nonident}. Prediction alone therefore does not determine a preferred coordinate system. Operator complexity supplies an additional inductive criterion: among predictively adequate coordinates, prefer those whose transition is easier to describe. This criterion does not guarantee that the selected coordinates coincide with physically meaningful variables.

\subsection{Dynamics compression creates a collapse shortcut}

\begin{proposition}[Degenerate minimum without representation constraints]
\label{prop:collapse}
Assume that the encoder classes contain a constant map
$E_0(x)=z_0$ and that the predictor class contains a transition $H_0$ satisfying
\begin{equation}
H_0(z_0,\epsinfo)
=
z_0
\end{equation}
for every admissible $\epsinfo$. Suppose further that $H_0$ attains the minimum value of
$\lambda_{\mathrm s}\Csym(\mathcal E)
+\lambda_{\mathrm c}\Rcorr(\phi)$
within the predictor class and that all objective terms are nonnegative. Then the unconstrained objective
\begin{equation}
\Lpred
+
\lambda_{\mathrm s}\Csym(\mathcal E)
+
\lambda_{\mathrm c}\Rcorr(\phi)
\end{equation}
admits the collapsed solution $(E_0,E_0,H_0)$ as a global minimiser whenever
$d(z_0,z_0)=0$.
\end{proposition}

This proposition says that, without explicit representation constraints, the easiest way to obtain simple and perfectly predictable latent dynamics may be to remove all variation from the representation. Once both encoders map every observation to the same point, any predictor for which $z_0$ is a fixed point achieves zero latent prediction error. For example, the identity transition
\begin{equation}
H_0(z,\epsinfo)
=
z
\end{equation}
satisfies
$H_0(z_0,\epsinfo)=z_0$
for every $z_0$. The model may therefore attain zero predictive loss and minimal operator complexity while retaining no information about the observations or their dynamics.

Appendix~\ref{app:proof-collapse} gives the proof. This failure mode is more specific than the general possibility of collapse in non-contrastive representation learning: operator compression creates a direct incentive to make the transition trivial by first making the representation trivial. Representation constraints are therefore essential to ensure that the method simplifies the governing law rather than erasing the state on which it acts.

\subsection{Controlling allocation to the correction}

\begin{proposition}[Non-identifiability of the hybrid decomposition]
\label{prop:decomp}
Let $r$ be any function for which $\Fsym+r$ remains in the symbolic function class and $c_\phi-r$ remains in the correction class. Then
\begin{equation}
\Fsym+c_\phi
=
(\Fsym+r)
+
(c_\phi-r),
\end{equation}
so predictive loss alone cannot determine how structure is allocated between the two components.
\end{proposition}

Intuitively, the same overall predictor can be obtained by transferring part of the transition function between the symbolic law and the neural correction.
This fact, proved in Appendix~\ref{app:proof-decomp}, motivates two forms of control. The symbolic-complexity penalty prevents uncontrolled symbolic growth, while $\Rcorr$ prevents a powerful correction network from explaining predictable structure unnecessarily. The resulting decomposition represents an inductive preference; stronger identifiability would require additional assumptions on the grammar, correction class, data support, or relationship between the two function spaces.

\begin{proposition}[Regularised allocation to the correction]
\label{prop:correction}
Fix the encoders and a symbolic law $F$. Let
\begin{equation}
m(z,\epsinfo)
=
\E[Z_T\mid Z_C=z,\epsinfo],
\end{equation}
and suppose the correction ranges over all square-integrable functions. For $\lambda>0$, the minimiser of
\begin{equation}
\label{eq:correction-allocation-objective}
\E
\left[
\left\|
Z_T
-
F(Z_C,\epsinfo)
-
c(Z_C,\epsinfo)
\right\|_2^2
+
\lambda
\left\|
c(Z_C,\epsinfo)
\right\|_2^2
\right]
\end{equation}
is
\begin{equation}
\label{eq:correction-shrinkage}
c^\star(z,\epsinfo)
=
\frac{
m(z,\epsinfo)-F(z,\epsinfo)
}{
1+\lambda
}
\end{equation}
almost surely.
\end{proposition}

The proof is given in Appendix~\ref{app:proof-correction}. Intuitively, the regulariser prevents the correction from fully absorbing the discrepancy left by the symbolic law, with larger $\lambda$ assigning less of that discrepancy to the correction.
When $\lambda=0$, an unrestricted correction absorbs the complete conditional-mean discrepancy $m-F$. For $\lambda>0$, this discrepancy is shrunk by the factor $(1+\lambda)^{-1}$, limiting how much predictive structure is assigned to the correction. Because Proposition~\ref{prop:correction} holds $F$ fixed, stronger regularisation leaves more discrepancy unexplained in this analytical setting. In the full joint optimisation, however, it encourages structure expressible by the symbolic grammar to be retained in the symbolic law. A finite neural correction generally learns a regularised projection of the residual onto its chosen function class.

\subsection{Long-horizon prediction is a separate requirement}

One-step adequacy does not imply accurate or non-divergent recursive prediction. Let $T$ denote the true latent transition and $H$ the learned hybrid transition.

\begin{proposition}[Rollout error under uniform one-step adequacy]
\label{prop:rollout}
Assume that $H(\cdot,u)$ is $L$-Lipschitz for every relevant action or conditioning value $u$, and that
\begin{equation}
\sup_{z,u}
\left\|
T(z,u)-H(z,u)
\right\|_2
\le
\delta.
\end{equation}
If
\begin{equation}
e_h
=
\left\|
z_{t+h}
-
\widehat z_{t+h}
\right\|_2,
\end{equation}
then
\begin{equation}
\label{eq:rollout-bound}
e_h
\le
L^h e_0
+
\begin{cases}
\displaystyle
\delta\frac{L^h-1}{L-1},
&
L\neq1,
\\[6pt]
h\delta,
&
L=1.
\end{cases}
\end{equation}
\end{proposition}

Intuitively, long-horizon error depends not only on the one-step approximation error $\delta$, but also on how strongly the learned transition amplifies existing errors through its Lipschitz constant $L$.
Appendix~\ref{app:proof-rollout} gives the proof. The bound clarifies both the promise and the limitation of operator compression: a compact law may reduce local approximation error and expose properties relevant to stability analysis, but compactness alone does not guarantee $L<1$ or prevent errors from accumulating. Long-horizon prediction must therefore be assessed separately through recursive rollout metrics, divergence rates, and, where appropriate, invariant or attractor statistics.

\section{Representation and Dynamics Learning}
\label{sec:representation-dynamics}

This section turns the constrained principle of Section~\ref{sec:elegant} into a practical learning procedure. We first describe the representation regulariser that preserves informative, non-collapsed latent states. We then introduce a unified practical objective combining prediction, rollout, representation quality, symbolic complexity, and correction control. Finally, we present two deployment modes: end-to-end alternating learning, which jointly adapts the coordinates and their dynamics, and frozen-encoder learning, which discovers dynamics in an existing representation space. Exact experimental instantiations are given in Section~\ref{sec:experiments}, while stage-specific objectives and optimisation details are deferred to Appendix~\ref{app:optimization}.

\subsection{A modular representation regulariser}

The representation module should preserve information needed for prediction while preventing the collapse shortcut identified in Proposition~\ref{prop:collapse}. We write its training objective abstractly as
\begin{equation}
\label{eq:repr-objective}
\mathcal L_{\mathrm{repr}}
=
\Lpred
+
\lambda_{\mathrm r}\Lroll
+
\lambda_{\mathrm{IB}}\RIB,
\end{equation}
where $\lambda_{\mathrm r}=0$ when multi-step prediction is not applicable.

One information-theoretic choice is the predictive bottleneck
\begin{equation}
\label{eq:ib}
\RIB^{\mathrm{MI}}
=
\I(X_C;Z_C)
-
\beta_{\mathrm{rel}}\I(Z_C;Z_T),
\end{equation}
which compresses the context representation while preserving information relevant to the target \citep{huang2026ibvjepa}. In practice, $\RIB$ may instead be implemented using a variational bottleneck, VICReg-style variance and covariance constraints, Barlow Twins, latent noise, dimensional restrictions, or related non-collapse mechanisms. These alternatives are not mathematically equivalent; they are modular ways of enforcing an informative predictive representation. Appendix~\ref{app:ibforms} describes representative choices.

Later, our experiments use a VICReg-style regulariser because its variance term directly prevents constant embeddings, while its covariance and invariance terms discourage redundant coordinates and encourage consistency under the selected observation augmentations. Its exact form is specified with the experimental objectives in Section~\ref{sec:experiments}.

\subsection{Unified practical objective}
\label{sec:practical-objectives}

A practical scalar relaxation of the constrained objective in Equation~\eqref{eq:constrained} is
\begin{equation}
\label{eq:lagrangian}
\mathcal L_{\ours}^{\mathrm{train}}
=
\Lpred
+
\lambda_{\mathrm r}\Lroll
+
\lambda_{\mathrm{IB}}\RIB
+
\lambda_{\mathrm s}
\Csym^{\mathrm{train}}(\mathcal E,\alpha)
+
\lambda_{\mathrm c}\Rcorr(\phi),
\end{equation}
where all weights are nonnegative. The five terms respectively control one-step prediction, recursive prediction, representation quality, symbolic parsimony, and reliance on the neural correction.

The optimisation-compatible complexity
$\Csym^{\mathrm{train}}$
may be the discrete structural score
$\Csym(\mathcal E)$
or a differentiable surrogate when symbolic coefficients and support are learned continuously. The final discovered law is evaluated using the thresholded structural complexity defined in Equation~\eqref{eq:omega}. Exact forms are given in Appendix~\ref{app:optimization}.

Different learning stages use restrictions of Equation~\eqref{eq:lagrangian}: terms that are constant or inapplicable in a particular stage are omitted. Because encoder learning and symbolic search are nonconvex, Equation~\eqref{eq:lagrangian} should be understood as a controllable scalarisation of the constrained principle rather than a guaranteed equivalent formulation. Bayesian fitting criteria and downstream planning objectives are separate from this deterministic training loss.

\subsection{End-to-end alternating learning}

End-to-end \ours jointly searches for an informative representation and a compact transition law. Direct joint optimisation is difficult because symbolic structure search is discrete or sparsity-driven, whereas encoder, coefficient, and correction learning are continuous. We therefore alternate between two complementary phases:

\begin{enumerate}[leftmargin=*,itemsep=1pt]
\item \textbf{Dynamics search:} hold the encoders fixed and find a compact symbolic law, together with a correction where required, in the current latent coordinates.
\item \textbf{Space search:} hold the symbolic structure fixed or locally relaxed and update the encoder and continuous dynamics parameters so that the representation remains predictive and non-collapsed while supporting a simpler transition.
\end{enumerate}

\begin{algorithm}[H]
\caption{Alternating training for end-to-end \ours}
\label{alg:alternating}
\begin{algorithmic}[1]
\REQUIRE Dataset $\D$, context encoder $E_\theta$, target encoder $E_{\bar\theta}$, grammar $\mathfrak E$, number of cycles $C$
\STATE Warm-start $(\theta,\bar\theta)$ using a neural JEPA predictor and the representation objective in Equation~\eqref{eq:repr-objective}
\STATE Initialise symbolic structure $\mathcal E$, coefficients $\alpha$, and correction $c_\phi$
\FOR{outer cycle $c=1,\ldots,C$}
    \STATE Encode latent tuples $\{(Z_C^{(i)},Z_T^{(i)},\epsinfo^{(i)})\}$
    \STATE \textbf{Dynamics search:} fix the encoders and minimise Equation~\eqref{eq:lagrangian} over $(\mathcal E,\alpha,\phi)$, omitting $\RIB$
    \STATE \textbf{Space search:} fix or locally relax $\mathcal E$ and minimise Equation~\eqref{eq:lagrangian} over $(\theta,\alpha,\phi)$
    \STATE Update $\bar\theta$ by exponential moving average of $\theta$
\ENDFOR
\RETURN $(E_\theta,E_{\bar\theta},F_{\mathcal E,\alpha},c_\phi)$
\end{algorithmic}
\end{algorithm}

Alternation implements the learned-transform interpretation of Section~\ref{sec:elegant}: improved coordinates permit simpler equations, while the symbolic objective provides pressure towards coordinates with lower induced-dynamics complexity. The representation regulariser prevents this pressure from being satisfied through collapse. Detailed phase-specific objectives, differentiable symbolic relaxations, and alternative optimisation schemes are given in Appendix~\ref{app:optimization}.

\subsection{Frozen pretrained encoders}

Representation learning is \textit{optional}. Given a frozen pretrained encoder $E_{\mathrm{pre}}$, such as a ViT, DINO, MAE, I-JEPA, or V-JEPA encoder \citep{dosovitskiy2021vit,caron2021dino,he2022mae,assran2023ijepa,bardes2024vjepa}, define
\begin{equation*}
Z_C
=
E_{\mathrm{pre}}(X_C),
\qquad
Z_T
=
E_{\mathrm{pre}}(X_T),
\end{equation*}
and optimise only $(\mathcal E,\alpha,\phi)$. Because the representation is fixed, $\RIB$ and the EMA update are omitted, while the symbolic and correction modules remain unchanged.
This mode turns
$\Cdyn(E_{\mathrm{pre}},E_{\mathrm{pre}})$
into a diagnostic of whether the pretrained representation already exposes compact dynamics. It is computationally simpler than end-to-end learning, but it cannot reshape coordinates that hide an otherwise simple transition.

\begin{table}[t]
\centering
\small
\caption{The two training modes share the same symbolic--correction transition model.}
\label{tab:modes}
\begin{tabularx}{\columnwidth}{p{0.19\columnwidth}p{0.27\columnwidth}X}
\toprule
Mode & Optimised modules & Main trade-off \\
\midrule
End-to-end
&
Encoders, symbolic law, and correction
&
Can search for coordinates with lower induced-dynamics complexity, but requires alternating optimisation and explicit non-collapse control.
\\
Frozen encoder
&
Symbolic law and correction
&
Reduces training cost and directly uses pretrained representations, but cannot repair coordinates that potentially conceal simple dynamics.
\\
\bottomrule
\end{tabularx}
\end{table}

\section{Bayesian \ours: Uncertainty over Latent Dynamics}

We further introduce uncertainty into dynamics learning through a Bayesian formulation of \ours. The encoders remain deterministic and may be trained jointly beforehand or kept frozen, while Bayesian inference is applied only to the latent transition model. In particular, uncertainty is represented over the symbolic structure, its coefficients, the transition covariance, and an optional Gaussian-process correction.

\subsection{Bayesian symbolic regression in latent space}

The deterministic SJEPA formulation selects a single symbolic transition law. With finite data, however, several expressions may explain the observed latent transitions nearly equally well, leaving uncertainty about the symbolic structure, its coefficients, and residual transition variability. Bayesian \ours represents this uncertainty explicitly rather than committing immediately to one equation. Throughout this section, the encoders are treated as fixed, so uncertainty is introduced \textit{only} over the latent dynamics.

Given latent transition data
\begin{equation*}
\D_Z
=
\left\{
\left(
z_C^{(i)},
z_T^{(i)},
\epsinfo^{(i)}
\right)
\right\}_{i=1}^n,
\end{equation*}
we model each target embedding as
\begin{equation}
\label{eq:bayes-like}
z_T^{(i)}
\mid
z_C^{(i)},
\epsinfo^{(i)},
\mathcal E,
\alpha,
\Sigma
\sim
\N \left(
F_{\mathcal E,\alpha}
\left(
z_C^{(i)},
\epsinfo^{(i)}
\right),
\Sigma
\right).
\end{equation}
Here, $\mathcal E$ specifies the symbolic structure, $\alpha$ contains its numerical coefficients, and $\Sigma$ represents latent transition variability not explained by the symbolic mean.

To favour parsimonious laws, we place the complexity prior
\begin{equation}
\label{eq:structure-prior}
p(\mathcal E)
=
\frac{
\exp \left\{
-\gamma\Csym(\mathcal E)
\right\}
}{
Z_\gamma
},
\qquad
Z_\gamma
=
\sum_{\mathcal E\in\mathfrak E}
\exp \left\{
-\gamma\Csym(\mathcal E)
\right\},
\end{equation}
where $\gamma\ge0$ controls the preference for simpler expressions. We assume that the candidate structure space is finite or countable and that $Z_\gamma<\infty$. Together with priors
$p(\alpha\mid\mathcal E)$
and
$p(\Sigma)$,
Bayes' rule gives
\begin{equation}
\label{eq:posterior}
p(\mathcal E,\alpha,\Sigma\mid\D_Z)
\propto
p(\D_Z\mid\mathcal E,\alpha,\Sigma)
p(\alpha\mid\mathcal E)
p(\Sigma)
p(\mathcal E).
\end{equation}
The posterior balances predictive fit, symbolic complexity, and prior plausibility of the coefficients and residual covariance.

For a new context embedding $z_C$ and side information $\epsinfo$, the posterior predictive distribution averages over plausible symbolic structures and parameter values:
\begin{equation}
\label{eq:postpred}
p(z_T\mid z_C,\epsinfo,\D_Z)
=
\sum_{\mathcal E}
\int
p(z_T\mid z_C,\epsinfo,\mathcal E,\alpha,\Sigma)
p(\mathcal E,\alpha,\Sigma\mid\D_Z)
\dd\alpha\,\dd\Sigma.
\end{equation}
This distribution propagates uncertainty about the symbolic law, its coefficients, and residual transition variability into the predicted target embedding. Bayesian \ours can therefore retain several competing explanations when the available data do not identify one law decisively.

\begin{theorem}[MAP equivalence for penalised symbolic regression]
\label{thm:map}
Assume that the latent transitions are conditionally independent under Equation~\eqref{eq:bayes-like}, with fixed isotropic covariance
$\Sigma=\sigma^2I$
for $\sigma^2>0$. Let
\begin{equation*}
p(\mathcal E,\alpha)
=
p(\alpha\mid\mathcal E)p(\mathcal E),
\end{equation*}
where $p(\mathcal E)$ is the proper complexity prior in Equation~\eqref{eq:structure-prior}. Then any maximum-a-posteriori estimator of $(\mathcal E,\alpha)$ minimises
\begin{equation}
\label{eq:map-objective}
\frac{1}{2\sigma^2}
\sum_{i=1}^n
\left\|
z_T^{(i)}
-
F_{\mathcal E,\alpha}
\left(
z_C^{(i)},
\epsinfo^{(i)}
\right)
\right\|_2^2
+
\gamma\Csym(\mathcal E)
-
\log p(\alpha\mid\mathcal E),
\end{equation}
and any minimiser of Equation~\eqref{eq:map-objective} is a MAP estimator, provided the extrema exist.
\end{theorem}

The proof is given in Appendix~\ref{app:proof-map}. Intuitively, the theorem shows that deterministic symbolic penalties can be interpreted as negative log-priors: penalising complex expressions corresponds to assigning them lower prior probability.
The symbolic-complexity penalty is induced by the structure prior, while coefficient regularisation is determined by
$p(\alpha\mid\mathcal E)$.
MAP estimation returns one most probable symbolic law and coefficient vector. By contrast, Equation~\eqref{eq:postpred} averages over competing structures and parameter values, retaining uncertainty when the data do not identify one law decisively.

\subsection{Bayesian hybrid predictor with a Gaussian-process correction}

To represent systematic residual dynamics not captured by the symbolic grammar, we augment the symbolic mean with a Gaussian-process correction\footnote{Equation~\eqref{eq:gp-hybrid} is the direct-transition formulation. For a continuous-time vector-field model observed at step size $\Delta t_t$, the corresponding one-step likelihood may instead use
\begin{equation*}
z_{t+1}
\mid
z_t,
\Delta t_t,
\mathcal E,
\alpha,
g,
\Sigma_{\Delta t_t}
\sim
\mathcal N \left(
z_t
+
\Delta t_t
\left[
F_{\mathcal E,\alpha}(z_t)
+
g(z_t)
\right],
\Sigma_{\Delta t_t}
\right).
\end{equation*}
Our later experiments use the deterministic counterpart of this residual formulation, whereas the task-general Bayesian development is written using the direct-transition notation.}:
\begin{equation}
\label{eq:gp-hybrid}
z_T
=
F_{\mathcal E,\alpha}(z_C,\epsinfo)
+
g(z_C,\epsinfo)
+
\eta,
\qquad
g\sim\mathcal{GP}(0,k),
\qquad
\eta\sim\mathcal N(0,\Sigma).
\end{equation}
The Gaussian process represents input-dependent residual structure, whereas $\eta$ represents transition variability remaining after conditioning on the symbolic law and correction. As in the deterministic hybrid model, the symbolic--GP decomposition is not identifiable from predictive fit alone. Its allocation depends on the symbolic-structure prior, coefficient priors, GP kernel family, kernel-amplitude prior, and transition-noise prior. These priors provide the Bayesian analogue of deterministic correction control: they encode a preference for a compact symbolic explanation with a modest residual correction, but they do not guarantee a unique decomposition.

For fixed $(\mathcal E,\alpha)$, define the GP input
\begin{equation*}
s_i
=
\left(
z_C^{(i)},
\epsinfo^{(i)}
\right).
\end{equation*}
, and the symbolic residual
$
r_i
=
z_T^{(i)}
-
F_{\mathcal E,\alpha}
\left(
z_C^{(i)},
\epsinfo^{(i)}
\right).
$
Stacking the residuals gives
\begin{equation}
\mathbf r
=
\operatorname{vec}
\left(
[r_1,\ldots,r_n]^\top
\right)
\in
\mathbb R^{nd_z},
\end{equation}
where $d_z$ is the latent dimension. Let $\mathbf K_g$ denote the covariance matrix induced by the vector-valued GP over the stacked inputs and outputs. After integrating out $g$, the residual vector has the marginal distribution
\begin{equation}
\label{eq:gp-marginal}
\mathbf r
\mid
\mathcal E,\alpha
\sim
\N \left(
0,
\mathbf C
\right),
\qquad
\mathbf C
=
\mathbf K_g
+
I_n\otimes\Sigma.
\end{equation}
Let $\vartheta_g$ denote the GP kernel hyperparameters. The corresponding negative log marginal likelihood is \citep{Carl2005GP}
\begin{equation}
\label{eq:gp-nlml_main}
-\log
p(\D_Z\mid\mathcal E,\alpha, \vartheta_g, \Sigma)
=
\underbrace{
\frac{1}{2}
\mathbf r^\top
\mathbf C^{-1}
\mathbf r
}_{\text{residual data fit}}
+
\underbrace{
\frac{1}{2}
\log|\mathbf C|
}_{\text{complexity penalty}}
+
\frac{nd_z}{2}\log(2\pi).
\end{equation}
The quadratic term measures the symbolic residual after accounting for the covariance structure permitted by the GP and transition noise. The log-determinant term is an Occam factor: it penalises covariance structures that can explain a broad range of residual functions, unless their additional flexibility is supported by improved fit. The final term is constant with respect to the model parameters.

When independent GPs are used for the latent coordinates, Equation~\eqref{eq:gp-nlml_main} decomposes across coordinates. Let $\mathbf r_j\in\mathbb R^n$ be the residuals for coordinate $j$ and let
\begin{equation}\tag{\ref{eq:gp-marginal}b}
\mathbf C_j
=
\mathbf K_j
+
\sigma_j^2 I_n.
\end{equation}
Then
\begin{equation} \tag{\ref{eq:gp-nlml_main}b}
-\log
p(\D_Z\mid\mathcal E,\alpha, \vartheta_g, \Sigma)
=
\sum_{j=1}^{d_z}
\left[
\frac{1}{2}
\mathbf r_j^\top
\mathbf C_j^{-1}
\mathbf r_j
+
\frac{1}{2}
\log|\mathbf C_j|
+
\frac{n}{2}\log(2\pi)
\right].
\end{equation}
The Bayesian hybrid predictor provides a probabilistic analogue of the symbolic--correction decomposition in deterministic \ours. The resulting allocation remains prior-dependent, while the marginal likelihood balances residual fit against the covariance flexibility permitted by the selected GP kernel and quantifies uncertainty about unexplained dynamics. 

\section{Explicit Action Dependence and Planning}

In control tasks, the side information $\epsinfo$ may contain an action $u_t$. For a discrete-time direct-transition model, the action-conditioned predictor is\footnote{For a continuous-time vector-field realisation, the planner instead uses the integrated transition
\begin{equation*}
\widehat z_{t+1}
=
\Hpredtt(z_t,u_t)
=
\mathcal I_{\Delta t_t}
\left(
z_t,
F_{\mathcal E,\alpha}+c_\phi,
u_t
\right).
\end{equation*}
}
\begin{equation}
\label{eq:action}
\widehat z_{t+1}
=
\Hpred(z_t,u_t)
=
F_{\mathcal E,\alpha}(z_t,u_t)
+
c_\phi(z_t,u_t).
\end{equation}
The symbolic component makes the role of the action explicit. For example, the discovered law may reveal whether an action affects a particular latent coordinate linearly, through an interaction with the current state, or only within a particular operating regime. This provides information that is difficult to extract directly from an opaque neural predictor.

When $\Hpred$ is differentiable, its behaviour around the current (latent) state--action pair $(z_t,u_t)$ can be approximated by a first-order local model. Define\footnote{For the forward-Euler realisation $
\Hpredt(z)
=
z
+
\Delta t
\left[
F_{\mathcal E,\alpha}(z)
+
c_\phi(z)
\right]
$, the state Jacobian of the complete transition is
\begin{equation*}
A_t
=
I
+
\Delta t_t
\left[
\frac{\partial F_{\mathcal E,\alpha}}{\partial z}
+
\frac{\partial c_\phi}{\partial z}
\right]_{(z_t,u_t)},
\end{equation*}
whereas the corresponding continuous-time local analysis uses the Jacobian of the vector field itself. These two stability criteria should not be conflated.}
\begin{equation}
\label{eq:linearization}
A_t
=
\left.
\frac{\partial \Hpred(z,u)}{\partial z}
\right|_{(z_t,u_t)},
\qquad
B_t
=
\left.
\frac{\partial \Hpred(z,u)}{\partial u}
\right|_{(z_t,u_t)}.
\end{equation}
For small perturbations $\Delta z_t$ and $\Delta u_t$, these matrices give
\begin{equation}
\label{eq:local-linear-model}
\Delta z_{t+1}
\approx
A_t\Delta z_t
+
B_t\Delta u_t.
\end{equation}
Intuitively, $A_t$ describes how a small change in the current latent state propagates to the next state, while $B_t$ describes how a small change in the action affects the next state. Thus, $A_t$ captures local state sensitivity and $B_t$ captures local action sensitivity.

These matrices provide a direct interface to classical local analysis. Around an equilibrium $(z^\star,u^\star)$ satisfying
\begin{equation}
z^\star
=
\Hpred(z^\star,u^\star),
\end{equation}
the eigenvalues of $A^\star$ can be used to assess local stability, while the pair $(A^\star,B^\star)$ can be examined to determine which latent directions are locally influenced by the available actions. The same local model may also be used by linear or locally linear control procedures, or as an approximation inside iterative trajectory-optimization and model-predictive-control methods. These linearised analyses do not guarantee global stability or controllability; they describe the model only in a neighbourhood of the selected operating point.

For polynomial or elementary symbolic expressions, the derivatives of $F_{\mathcal E,\alpha}$ can often be obtained analytically. Automatic differentiation can be used for the neural correction $c_\phi$, so that
\begin{equation}
A_t
=
\left.
\frac{\partial F_{\mathcal E,\alpha}}{\partial z}
\right|_{(z_t,u_t)}
+
\left.
\frac{\partial c_\phi}{\partial z}
\right|_{(z_t,u_t)},
\qquad
B_t
=
\left.
\frac{\partial F_{\mathcal E,\alpha}}{\partial u}
\right|_{(z_t,u_t)}
+
\left.
\frac{\partial c_\phi}{\partial u}
\right|_{(z_t,u_t)}.
\end{equation}
This decomposition further shows whether the dominant action sensitivity is explained by the symbolic law or delegated to the neural correction.

Beyond local analysis, a generic external planner can use the full nonlinear hybrid predictor to simulate candidate action sequences. Let $K$ denote the planning horizon. Starting from the current latent state $\widehat z_t=z_t$, the planner recursively computes
\begin{equation}
\label{eq:planning-rollout}
\widehat z_{t+h+1}
=
\Hpred(
\widehat z_{t+h},
u_{t+h}
),
\qquad
h=0,\ldots,K-1.
\end{equation}
It then selects an action sequence by solving
\begin{equation}
\label{eq:planning}
u^\star_{t:t+K-1}
=
\argmin_{u_{t:t+K-1}}
\left[
\sum_{h=0}^{K-1}
\ell(
\widehat z_{t+h},
u_{t+h}
)
+
\ell_T(
\widehat z_{t+K}
)
\right],
\end{equation}
where $\ell$ is a stage cost and $\ell_T$ is a terminal cost.

The planner itself is not the methodological contribution of \ours. The contribution is an action-conditioned predictive model whose governing structure can be inspected, differentiated, locally linearised, and used by a range of external planning or control methods. In Bayesian \ours, the deterministic rollout can be replaced by a posterior predictive rollout, and the planner may minimise posterior expected cost or a risk-sensitive objective that accounts for uncertainty.

Compact action-conditioned laws may also improve sample efficiency and extrapolation by reusing the same structural mechanism across states and action sequences. This remains an empirical hypothesis rather than an unconditional guarantee: a misspecified symbolic grammar, an inaccurate latent representation, or a dominant neural correction can reduce both structural clarity and planning performance.

\section{Experiments}
\label{sec:experiments}

We conduct two controlled experiments targeting the two claims that define \ours. Experiment~1 asks \textit{whether joint representation and operator learning discovers predictive coordinates with simpler induced dynamics than post-hoc equation discovery}. Experiment~2 removes representation learning and tests \textit{whether correction regularisation preserves a compact symbolic mechanism under controlled grammar misspecification}. Detailed data generation, architectures, optimisation settings, checkpoint selection, per-seed equations, and additional metrics are given in Appendix~\ref{app:experimental-details}. Unless stated otherwise, reported values are means and standard deviations over three optimisation seeds.

\paragraph{Residual dynamics and evaluation protocol.}
Both experiments instantiate Equation~\eqref{eq:hybrid} using residual vector-field predictors,
\begin{equation}
\label{eq:experimental-residual-predictor}
\widehat z_{t+1}
=
z_t
+
\Delta t_t
\left[
F_{\mathcal E,\alpha}(z_t)
+
c_\phi(z_t)
\right],
\end{equation}
with $c_\phi\equiv0$ for symbolic-only models. This parameterisation treats
$F_{\mathcal E,\alpha}(z_t)+c_\phi(z_t)$
as an estimate of the continuous-time latent derivative $\dot z_t$ and obtains the next state through a forward-Euler step, while explicitly accounting for variable step sizes $\Delta t_t$. Training combines one-step prediction with a ten-step free rollout. Training and model selection use disjoint trajectory splits. Model parameters and symbolic coefficients are learned from the training trajectories, while separate validation trajectories are used to select checkpoints and, in Experiment~1, the best completed alternating cycle. When validation risks are nearly tied, the lower-complexity thresholded symbolic model is preferred. Test and OOD trajectories are used only for final evaluation after all model selection is complete. Experiment~1 evaluates prediction directly in the latent space and evaluates physical-state rollouts after fitting an affine map from training latents to the physical state\footnote{The affine map converts learned latent coordinates into physical-state coordinates so that rollout errors can be measured meaningfully.}; Experiment~2 directly uses the physical state as the latent coordinate, $z_t=(q_t,p_t)$, so predicted rollouts can be evaluated in physical-state space without fitting an additional affine alignment map.

\paragraph{Experimental objectives and model selection.}

Each experimental condition uses the applicable terms from the unified objective in Equation~\eqref{eq:lagrangian}; Table~\ref{tab:experimental-objectives} summarises the active objective terms in each condition. Experiment~1 jointly studies representation and operator learning, whereas Experiment~2 fixes the state coordinates and isolates the allocation between the symbolic law and neural correction.

In Experiment~1, the representation regulariser is
\begin{equation}
\label{eq:experimental-rib}
\RIB^{\mathrm{exp}}
=
\lambda_{\mathrm{inv}}\mathcal L_{\mathrm{inv}}
+
\lambda_{\mathrm{var}}\mathcal L_{\mathrm{var}}
+
\lambda_{\mathrm{cov}}\mathcal L_{\mathrm{cov}}
+
\lambda_{\mathrm{mean}}\mathcal L_{\mathrm{mean}}.
\end{equation}
Here, $\mathcal L_{\mathrm{inv}}$ encourages consistency under observation augmentation, $\mathcal L_{\mathrm{var}}$ imposes a per-coordinate variance floor, $\mathcal L_{\mathrm{cov}}$ penalises off-diagonal latent covariance, and $\mathcal L_{\mathrm{mean}}$ provides weak centering. The variance term provides the principal protection against the constant-representation shortcut in Proposition~\ref{prop:collapse}. The weights in Equation~\eqref{eq:experimental-rib} control the relative contributions within $\RIB^{\mathrm{exp}}$, while $\lambda_{\mathrm{IB}}$ in Equation~\eqref{eq:lagrangian} controls its overall strength.

\begin{table*}[t]
\centering
\small
\setlength{\tabcolsep}{5pt}
\caption{Active terms from Equation~\eqref{eq:lagrangian} in each experimental condition. A dash indicates that the corresponding term is absent. The unregularised hybrid retains a neural correction but sets $\lambda_{\mathrm c}=0$, so the correction penalty is inactive.}
\label{tab:experimental-objectives}
\begin{tabular}{lccccc}
\toprule
Training condition
&
$\Lpred$
&
$\Lroll$
&
$\RIB$
&
$\Csym^{\mathrm{train}}$
&
$\Rcorr$
\\
\midrule
Experiment 1: Neural JEPA and warm start
&
\checkmark
&
\checkmark
&
\checkmark
&
--
&
--
\\
Experiment 1: post-hoc symbolic
&
\checkmark
&
\checkmark
&
--
&
\checkmark
&
--
\\
Experiment 1: \ours dynamics phase
&
\checkmark
&
\checkmark
&
--
&
\checkmark
&
--
\\
Experiment 1: \ours space phase
&
\checkmark
&
\checkmark
&
\checkmark
&
\checkmark
&
--
\\
Experiment 1: one-step collapse diagnostic
&
\checkmark
&
--
&
--
&
\checkmark
&
--
\\
Experiment 2: symbolic-only models
&
\checkmark
&
\checkmark
&
--
&
\checkmark
&
--
\\
Experiment 2: regularised hybrid
&
\checkmark
&
\checkmark
&
--
&
\checkmark
&
\checkmark
\\
Experiment 2: unregularised hybrid
&
\checkmark
&
\checkmark
&
--
&
\checkmark
&
--
\\
Experiment 2: neural-only dynamics
&
\checkmark
&
\checkmark
&
--
&
--
&
--
\\
\bottomrule
\end{tabular}
\end{table*}

Model parameters are learned from the training trajectories. Checkpoints and, for \ours, completed alternating cycles are selected using the held-out validation risk
\begin{equation}
\label{eq:validation-risk}
\mathcal R_{\mathrm{val}}
=
\Lpred^{\mathrm{val}}
+
\lambda_{\mathrm r}
\Lroll^{\mathrm{val}},
\end{equation}
where $\lambda_{\mathrm r}$ is the rollout weight used by the corresponding condition. 
For Experiment~1 conditions that update the encoder, candidates must satisfy the prescribed non-collapse diagnostics; the no-$\RIB$ and fixed-point diagnostic conditions are exempt from this requirement. Among candidates whose validation risks lie within a predefined relative tolerance, the lower-complexity symbolic law is selected. Full loss definitions, weights, diagnostics, and selection tolerances are given in Appendix~\ref{app:experimental-details}.

\subsection{Experiment 1: Learning coordinates with elegant dynamics}

\paragraph{Design.}
We simulate the continuous-time pendulum
\begin{equation}
\label{eq:exp1-true-dynamics}
\dot q
=
p,
\qquad
\dot p
=
-\sin(q),
\end{equation}
using fourth-order Runge--Kutta integration. At each transition, the time step is sampled independently from
\begin{equation*}
\Delta t_t
\in
\{0.025,0.04,0.06\}.
\end{equation*}

The two-dimensional physical state is embedded in a noisy $32$-dimensional observation by combining raw and nonlinear state features and applying fixed orthogonal mixing. Thus, $q$ and $p$ are not supplied as named input coordinates, although they remain linearly recoverable from the noiseless full observation because the raw state features are included. The experiment therefore tests predictive-coordinate selection under high-dimensional mixing and nonlinear distractor features rather than recovery from a genuinely nonlinear observation inverse. The context encoder and its EMA target copy map these observations to a common two-dimensional latent space.

We compare:
\begin{enumerate}[leftmargin=*,itemsep=1pt]
\item \textbf{Neural JEPA}, which jointly learns the encoder and a neural residual vector field;
\item \textbf{post-hoc symbolic}, which freezes the Neural JEPA coordinates and subsequently fits a symbolic vector field, as detailed in Appendix~\ref{app:posthoc-symbolic};
\item \textbf{\ours symbolic}, which alternates symbolic-dynamics search and representation-space search with $c_\phi\equiv0$;
\item an \textbf{unconstrained one-step collapse diagnostic}, which removes representation regularisation and rollout supervision and trains the symbolic model using only the one-step objective;
\item a \textbf{collapsed fixed-point control}, in which both encoders are initialised as the same nonzero constant map.
\end{enumerate}

The unconstrained one-step diagnostic sets both the representation-regularisation weight and rollout-loss weight to zero in order to expose the one-step fixed-point shortcut directly. It therefore demonstrates that the unconstrained one-step objective can reach the collapse mechanism identified in Proposition~\ref{prop:collapse}; it is not a matched ablation isolating the effect of removing $\RIB$ alone. The collapsed fixed-point control uses the same one-step objective and constructs the degenerate solution explicitly. Because a constant representation is preserved under repeated identity transitions, the existence of the shortcut itself is not specific to one-step training.

The symbolic library is
\begin{equation}
\Theta(z)
=
\left\{
1,z_1,z_2,
\sin(z_1),\sin(z_2),
\cos(z_1),\cos(z_2),
z_1^2,z_2^2,z_1z_2
\right\}.
\end{equation}
The reported symbolic complexity is the weighted complexity of the terms whose fitted coefficients exceed the fixed reporting threshold specified in Appendix~\ref{app:experimental-details}.

\paragraph{Joint coordinate search exposes a substantially simpler law.}
Table~\ref{tab:exp1-main} compares the 3 predictive models. Neural JEPA achieves the lowest raw prediction and rollout errors, as expected from its flexible neural vector field. The primary operator-compression comparison is therefore between post-hoc symbolic regression and joint symbolic \ours.

\begin{figure*}[t]
\centering
\begin{minipage}[t]{0.49\textwidth}
\centering
\includegraphics[width=\linewidth]{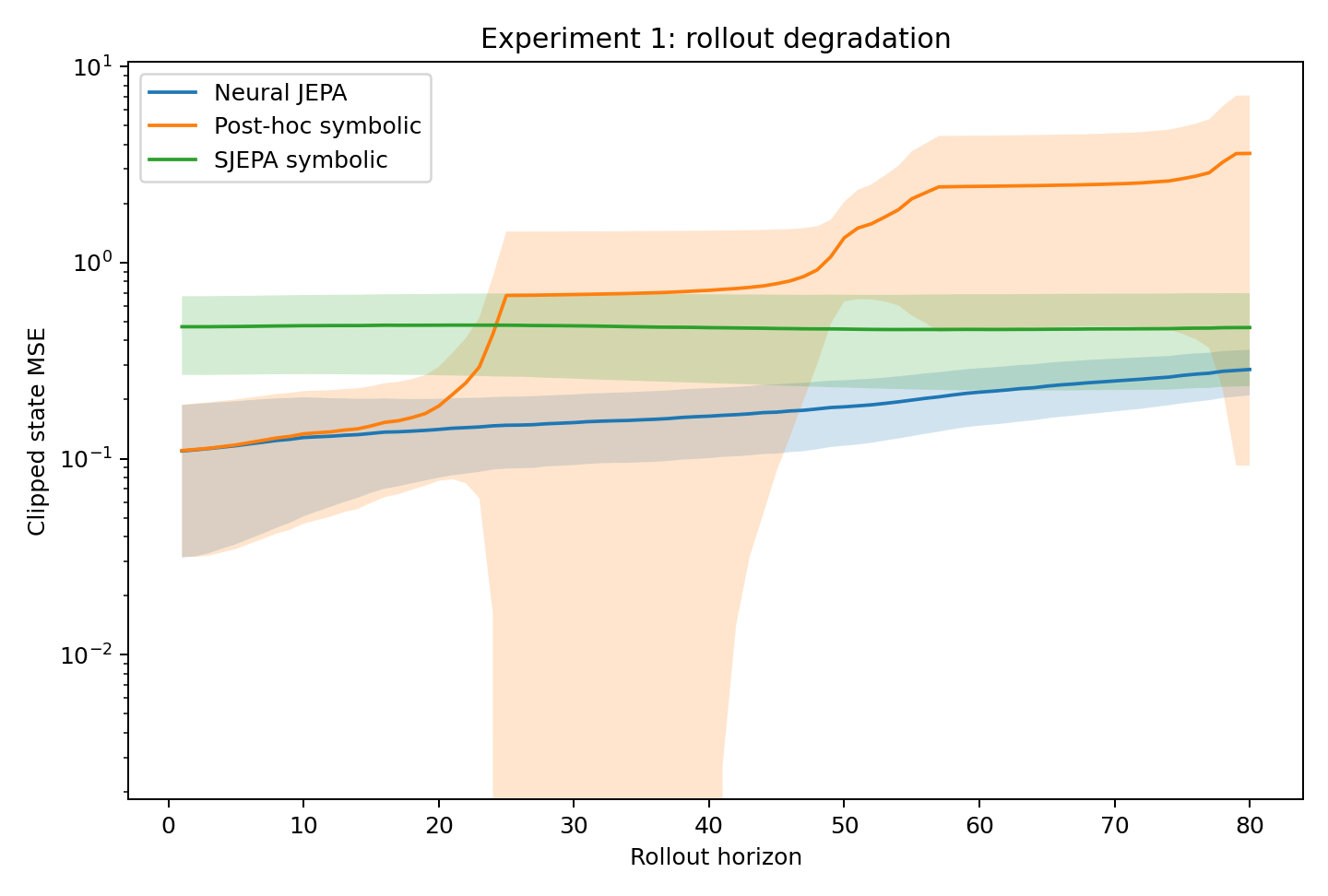}
\vspace{1mm}

\text{(a) Physical-state rollout error}
\end{minipage}
\hfill
\begin{minipage}[t]{0.49\textwidth}
\centering
\includegraphics[width=\linewidth]{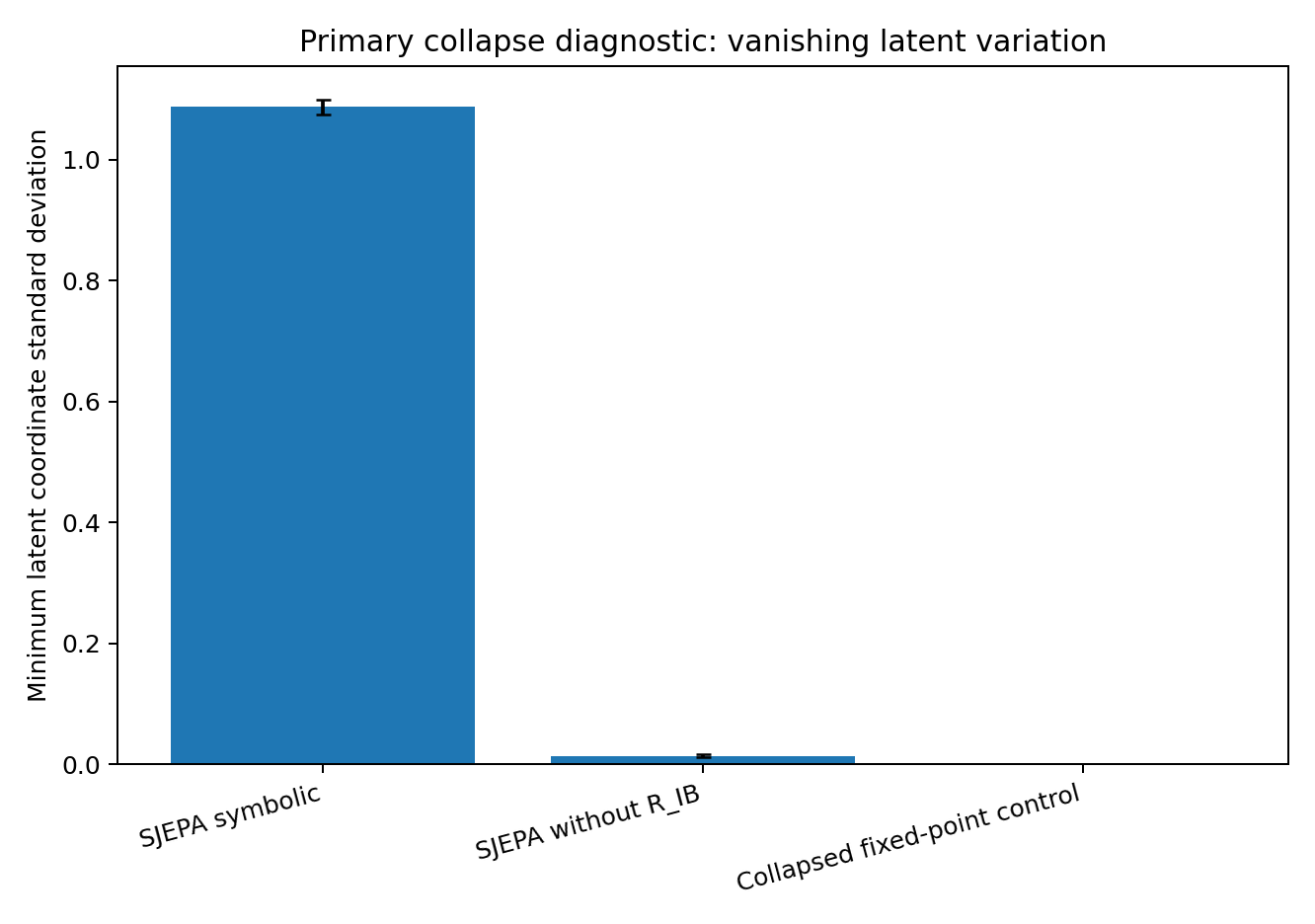}
\vspace{1mm}

\text{(b) Minimum latent coordinate standard deviation}
\end{minipage}
\caption{
Experiment~1 evaluates operator compression and representation collapse.
\textbf{(a)} Joint symbolic \ours accumulates substantially less physical-state rollout error than symbolic regression fitted post hoc, although the flexible Neural JEPA predictor remains the most accurate.
\textbf{(b)} Removing $\RIB$ causes absolute latent variation to nearly vanish, while the fixed-point control is effectively constant. Curves and error bars show means and one standard deviation over 3 seeds.
}
\label{fig:exp1-main}
\end{figure*}

Joint coordinate search reduces mean symbolic complexity from $26.0$ to $4.67$, a factor of approximately $5.6$. Within the respective learned latent spaces, it also reduces one-step latent MSE from
$2.09\times10^{-3}$
to
$5.81\times10^{-4}$,
a factor of approximately $3.6$. Because latent MSE depends on the scale and coordinate system of the learned representation, this comparison is descriptive rather than coordinate-invariant. The physical-state rollout results provide a complementary cross-model comparison: joint symbolic \ours reduces test rollout MSE from $1.229$ to $0.467$, OOD rollout MSE from $9.262$ to $3.435$, and OOD divergence rate from $0.178$ to $0.017$. Figure~\ref{fig:exp1-main}(a) shows the corresponding reduction in recursively accumulated error.

\begin{table*}[t]
\centering
\scriptsize
\setlength{\tabcolsep}{3.5pt}
\renewcommand{\arraystretch}{1.08}
\caption{Experiment~1 results. Latent MSE is evaluated within each model's learned embedding space and is therefore coordinate-dependent. Rollout columns report clipped physical-state MSE after applying an affine map fitted only on training latents; divergence thresholds are defined in Appendix~\ref{app:experimental-details}. A dash denotes a metric that is not applicable, and bold identifies the better of the two symbolic methods. Values are mean $\pm$ standard deviation over 3 optimisation seeds.}
\label{tab:exp1-main}
\begin{tabular}{lccccc}
\toprule
Method
&
Latent MSE $\downarrow$
&
$\Csym\downarrow$
&
Test rollout $\downarrow$
&
OOD rollout $\downarrow$
&
OOD divergence $\downarrow$
\\
\midrule
Neural JEPA
&
$9.81{\times}10^{-5}\pm3.93{\times}10^{-5}$
&
--
&
$0.180\pm0.066$
&
$2.376\pm1.433$
&
$0.000\pm0.000$
\\
Post-hoc symbolic
&
$2.09{\times}10^{-3}\pm2.48{\times}10^{-3}$
&
$26.00\pm3.24$
&
$1.229\pm0.709$
&
$9.262\pm4.021$
&
$0.178\pm0.143$
\\
\ours symbolic
&
$\mathbf{5.81{\times}10^{-4}\pm4.39{\times}10^{-4}}$
&
$\mathbf{4.67\pm2.49}$
&
$\mathbf{0.467\pm0.222}$
&
$\mathbf{3.435\pm1.394}$
&
$\mathbf{0.017\pm0.024}$
\\
\bottomrule
\end{tabular}
\end{table*}

The dominant \ours structure is consistent across seeds: every discovered law contains $z_2$ in the equation for $\dot z_1$ and $z_1$ in the equation for $\dot z_2$. Up to a reversal of latent orientation, the learned dynamics are therefore oscillator-like:
\begin{equation}
\label{eq:exp1-oscillator-summary}
\dot z_1
\approx
-a z_2,
\qquad
\dot z_2
\approx
b z_1
+
\text{lower-magnitude terms},
\qquad
a,b>0.
\end{equation}

To illustrate the difference between the three predictors, consider the selected models for the same seed. Neural JEPA represents the vector field by an unrestricted neural network,
\begin{equation}
\dot z
=
f_\psi(z),
\end{equation}
and therefore does not produce a closed-form symbolic equation. Symbolic regression fitted post hoc to the Neural JEPA coordinates gives
\begin{align*}
\dot z_1
&=
-1.189\sin(z_2)
-0.602
+0.414\cos(z_2)
+0.244\sin(z_1)
+0.238z_2^2
\nonumber\\
&\hspace{1.2cm}
-0.188z_1
+0.074\cos(z_1)
+0.069z_1^2,
\\
\dot z_2
&=
1.888\sin(z_1)
+1.133\cos(z_1)
-0.967
-0.386z_1
+0.320z_1^2
-0.245z_1z_2.
\end{align*}
By contrast, joint representation and equation learning produces
\begin{equation*}
\dot z_1
=
-0.806z_2,
\qquad
\dot z_2
=
0.822z_1
+
0.177
-
0.143z_1^2.
\end{equation*}
The complete per-seed \ours equations are reported in Appendix~\ref{app:exp1-equations}.

The post-hoc and Neural JEPA models share the same prediction-only coordinates, whereas \ours learns a different latent coordinate system. Their coefficients should therefore not be compared term by term. The relevant comparison is whether each representation admits an accurate and compact induced transition. In this example, the prediction-only coordinates require a mixture of polynomial and trigonometric terms, while joint representation and equation learning exposes a much more concise cross-coupled system.

Together with Table~\ref{tab:exp1-main}, these results support our operator-compression claim: \ours discovers substantially simpler symbolic dynamics and achieves lower physical-state rollout error than symbolic regression fitted post hoc. They do not imply that the symbolic model universally exceeds Neural JEPA in predictive accuracy. Rather, \ours trades some predictive flexibility for a considerably more concise and inspectable transition.

\paragraph{Representation constraints prevent trivial operator compression.}
The unconstrained one-step diagnostic\footnote{As this diagnostic removes both $\RIB$ and rollout supervision, it demonstrates that the unconstrained one-step objective can reach the theoretical fixed-point shortcut; it is not a matched ablation isolating the effect of removing $\RIB$ alone.} exhibits the degenerate behaviour predicted by Proposition~\ref{prop:collapse}. In all 3 seeds, the learned vector field becomes
\begin{equation*}
\dot z_1
=
0,
\qquad
\dot z_2
=
0.
\end{equation*}
Under the residual parameterisation in Equation~\eqref{eq:experimental-residual-predictor}, this zero vector field implements the identity transition
$\widehat z_{t+1}=z_t$.
Its latent prediction error is smaller than that of every non-collapsed model, but Table~\ref{tab:exp1-collapse} shows that this apparent success is obtained by reducing the representation to a region with almost no variation. The constructive fixed-point control reaches essentially exact collapse at a nonzero constant latent state.

Table~\ref{tab:exp1-main} evaluates the predictive and rollout performance of the three non-collapsed models. Table~\ref{tab:exp1-collapse} instead isolates the collapse mechanism by comparing regularised \ours with two diagnostic conditions. The no-$\RIB$ model tests whether collapse emerges when representation control is removed, while the fixed-point control constructs the constant-representation solution directly. We report absolute latent-variation statistics because a low latent prediction error can be obtained trivially when all observations are encoded near the same point.

\begin{table*}[t]
\centering
\scriptsize
\setlength{\tabcolsep}{7pt}
\caption{Experiment~1 representation-collapse diagnostics. Latent MSE is evaluated on test transitions, while $\min_j\operatorname{Std}(Z_j)$ and $\operatorname{Tr}(\operatorname{Cov}Z)$ are computed from context embeddings of the training trajectories. Regularised \ours is included as a non-collapsed reference; the no-$\RIB$ and fixed-point conditions are collapse diagnostics rather than matched predictive baselines. Very low latent prediction error is uninformative when absolute latent variation vanishes. Values are mean $\pm$ standard deviation over 3 optimisation seeds.}
\label{tab:exp1-collapse}
\begin{tabular}{lcccc}
\toprule
Method
&
Latent MSE $\downarrow$
&
$\min_j\operatorname{Std}(Z_j)$
&
$\operatorname{Tr}(\operatorname{Cov}Z)$
&
$\Csym$
\\
\midrule
\ours symbolic
&
$5.81{\times}10^{-4}\pm4.39{\times}10^{-4}$
&
$1.087\pm0.012$
&
$2.406\pm0.047$
&
$4.67\pm2.49$
\\
One-step collapse diagnostic
&
$4.15{\times}10^{-7}\pm1.78{\times}10^{-7}$
&
$0.0137\pm0.0023$
&
$(6.53\pm1.82){\times}10^{-4}$
&
$0$
\\
Collapsed fixed-point control
&
$(7.00\pm0.88){\times}10^{-10}$
&
$(3.69\pm0.09){\times}10^{-5}$
&
$<10^{-8}$
&
$0$
\\
\bottomrule
\end{tabular}
\end{table*}

Figure~\ref{fig:exp1-main}(b) visualises the corresponding loss of latent variation. Relative to the regularised model, the one-step collapse diagnostic reduces the minimum coordinate standard deviation by approximately two orders of magnitude, while the fixed-point control is effectively constant. The experiment therefore distinguishes genuine operator compression from representation collapse: the regularised model retains substantial two-dimensional variation, whereas the unconstrained objective obtains a trivial identity transition by making successive embeddings almost constant. Effective rank alone is insufficient for this diagnosis because it measures relative dimensional usage but is insensitive to the absolute scale of latent variation.

As reported in Table~\ref{tab:exp1-additional} (Appendix.\ref{app:experimental-details}), the learned \ours coordinates are less affinely aligned with the original physical state than the prediction-only coordinates: test affine-probe $R^2$ decreases from $0.814\pm0.063$ to $0.293\pm0.154$.
We therefore interpret Experiment~1 as evidence that joint learning can produce simpler, non-collapsed latent dynamics, rather than as evidence that operator compression preserves a simple affine correspondence with the original state. An illustrative example of the nonlinear latent geometry learned in Experiment~1 is shown in Figure~\ref{fig:additional-diagnostics}(a).

\subsection{Experiment 2: Hybrid dynamics under controlled grammar mismatch}

\paragraph{Design.}
Experiment~2 isolates the transition decomposition by using state-aligned coordinates, namely the physical state,
\begin{equation}
z
=
(q,p),
\end{equation}
and learning no encoder. The true dynamics include quadratic drag:
\begin{equation}
\label{eq:exp2-true-dynamics}
\dot q
=
p,
\qquad
\dot p
=
-\sin(q)-0.4p|p|.
\end{equation}
The incomplete output-specific grammar permits
\begin{equation}
\dot q:\{p\},
\qquad
\dot p:\{\sin(q)\},
\end{equation}
whereas the complete grammar additionally permits $p|p|$ in the second equation. Linear $p$ is deliberately excluded from $\dot p$ so that it cannot act as a symbolic surrogate for the omitted quadratic drag.

We compare symbolic-only dynamics under the incomplete and complete grammars, a neural-only vector field, and hybrid models with and without the correction penalty. The two hybrid models use the incomplete grammar and the same bounded neural correction architecture. Thus, their only methodological difference is whether $\Rcorr$ is active.

We measure reliance on the neural correction using the normalised correction-energy ratio
\begin{equation}
\label{eq:experimental-correction-share}
\tag{Eq.\ref{eq:correction-energy-ratio-appendix_vec}}
\rho_{\mathrm{corr}}
=
\frac{
\E \left[\|c_\phi(z)\|_2^2\right]
}{
\E \left[
\|F_{\mathcal E,\alpha}(z)+c_\phi(z)\|_2^2
\right]
+
\eta_0
},
\qquad
\eta_0=10^{-12}.
\end{equation}
A small value indicates limited reliance on the correction, but is meaningful only when considered jointly with predictive error and symbolic complexity. For symbolic-only models, $c_\phi\equiv0$ and hence $\rho_{\mathrm{corr}}=0$; the ratio is not applicable to the neural-only model because it has no symbolic--correction decomposition.

\begin{figure*}[t]
\centering
\begin{minipage}[t]{0.54\textwidth}
\centering
\includegraphics[width=\linewidth]{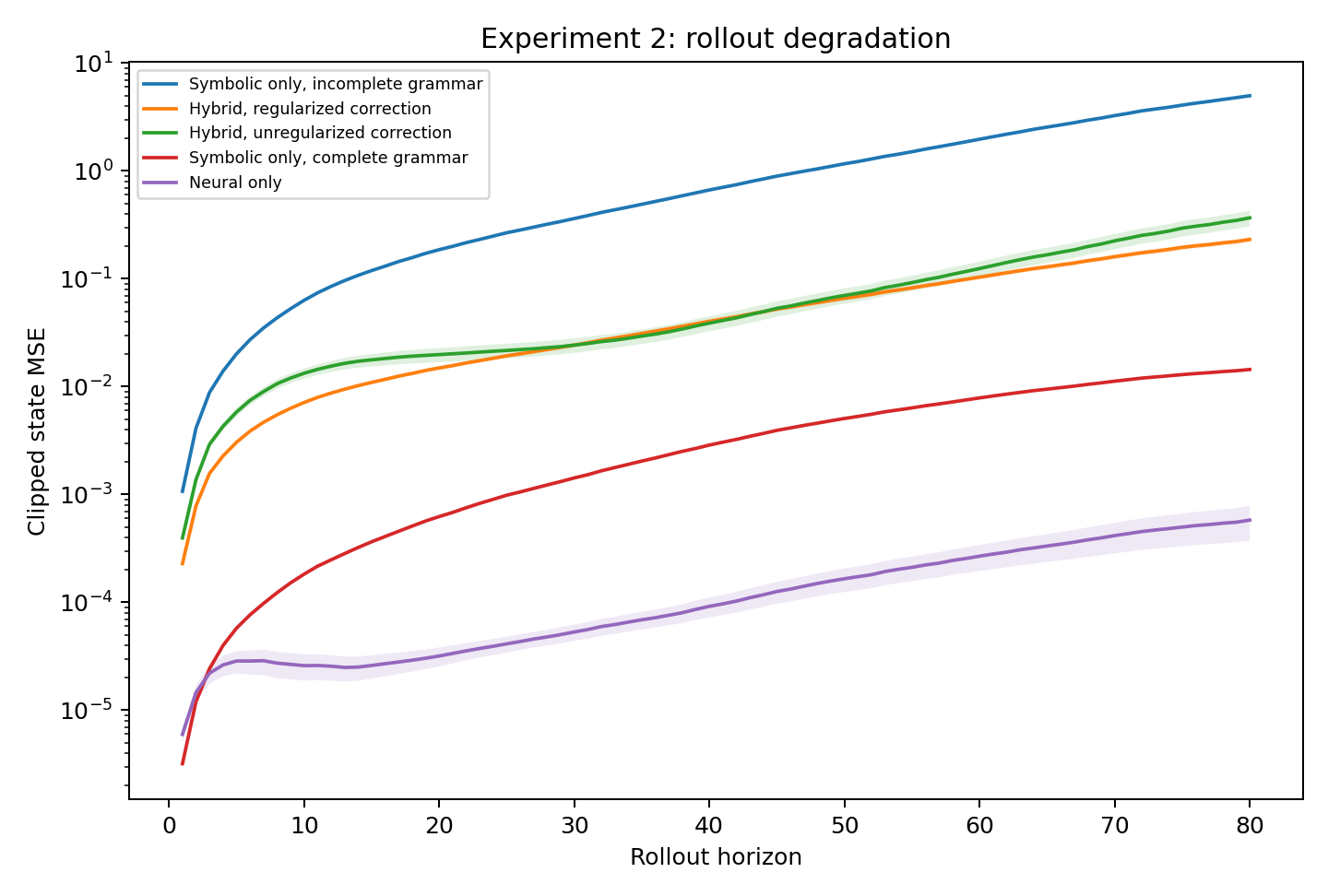}
\vspace{1mm}

\text{(a) State-space rollout error}
\end{minipage}
\hfill
\begin{minipage}[t]{0.45\textwidth}
\centering
\includegraphics[width=\linewidth]{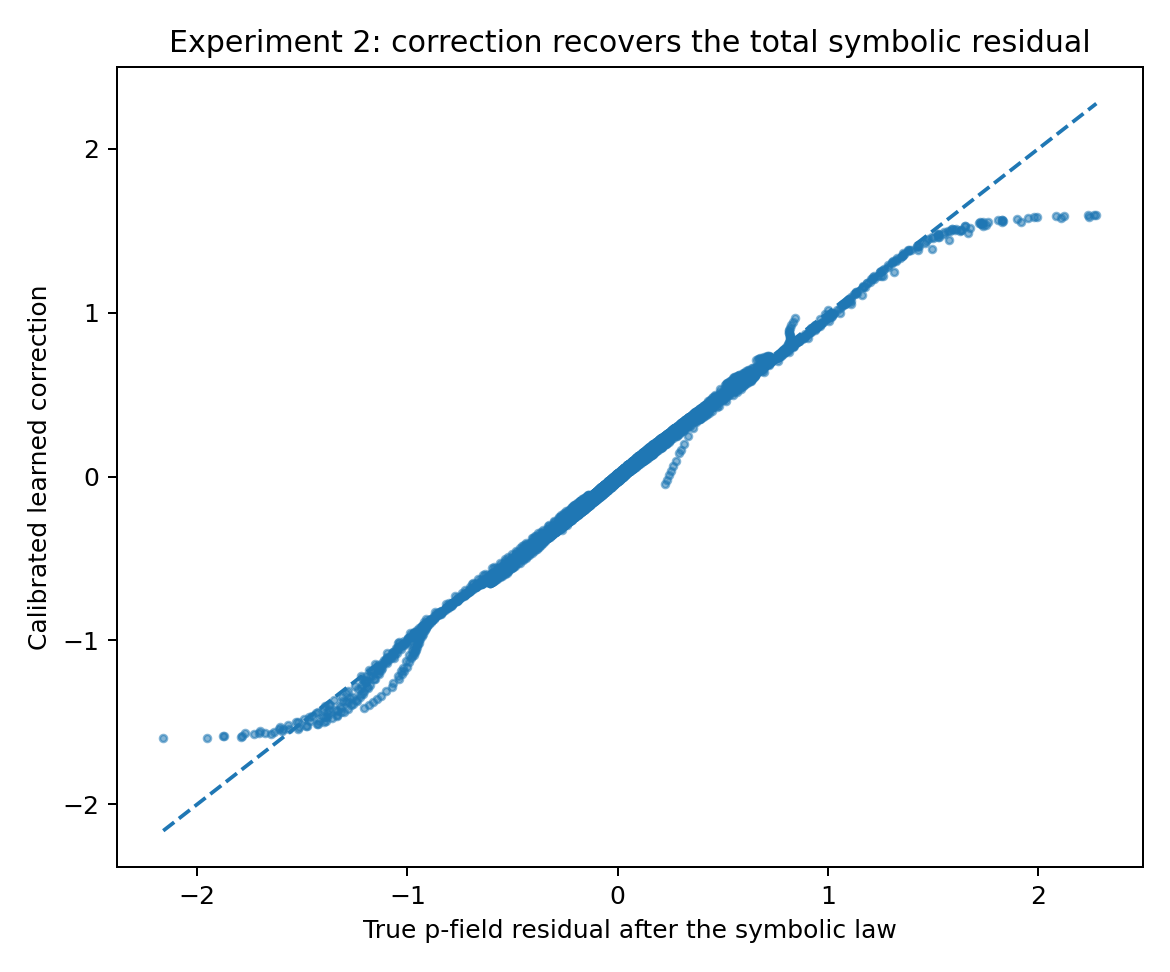}
\vspace{1mm}

\text{(b) Recovery of the symbolic residual}
\end{minipage}
\caption{
Experiment~2 evaluates controlled grammar mismatch and symbolic--neural allocation.
\textbf{(a)} The incomplete symbolic grammar accumulates substantial rollout error, whereas the complete symbolic grammar approaches the neural predictor. Both hybrid models substantially improve over the incomplete symbolic model.
\textbf{(b)} For the regularised hybrid using the \textit{incomplete} grammar, the neural correction closely tracks the held-out $p$-field residual left by the fitted symbolic component after scalar calibration on the training set. The dashed line denotes exact recovery.
}
\label{fig:exp2-main}
\end{figure*}

\paragraph{A complete grammar recovers the governing law.}
Table~\ref{tab:exp2-main} shows that the incomplete symbolic model underfits the drag mechanism, whereas restoring the omitted primitive reduces one-step state MSE by approximately two orders of magnitude and test rollout MSE by approximately $277$ times. Across all 3 seeds, the complete symbolic model selects exactly the intended three terms and recovers
\begin{equation*}
\label{eq:exp2-complete-recovery}
\dot q
=
(0.9830\pm0.0005)p,
\qquad
\dot p
=
-(0.9553\pm0.0010)\sin(q)
-
(0.3724\pm0.0007)p|p|.
\end{equation*}
The true coefficients are $(1,-1,-0.4)$, corresponding to relative errors of approximately $1.7\%$, $4.5\%$, and $6.9\%$. Figure~\ref{fig:exp2-main}(a) shows the corresponding long-horizon behaviour: the incomplete symbolic law accumulates substantial error, whereas restoring the omitted primitive produces rollouts approaching those of the neural predictor while retaining an explicit governing equation.

\begin{table*}[t]
\centering
\scriptsize
\setlength{\tabcolsep}{4pt}
\caption{Experiment~2 results under controlled grammar mismatch. The normalised correction-energy ratio $\rho_{\mathrm{corr}}$ measures reliance on the neural correction and should be interpreted jointly with predictive error and symbolic complexity. Values are mean $\pm$ standard deviation over three optimisation seeds.}
\label{tab:exp2-main}
\begin{tabular}{lcccccc}
\toprule
Method
&
State MSE $\downarrow$
&
Field MSE $\downarrow$
&
$\Csym\downarrow$
&
$\rho_{\mathrm{corr}}$
&
Test rollout $\downarrow$
&
OOD rollout $\downarrow$
\\
\midrule
Symbolic, incomplete
&
$(1.630\pm0.001){\times}10^{-4}$
&
$0.0835\pm0.00004$
&
$3$
&
$0$
&
$1.280\pm0.008$
&
$1.274\pm0.006$
\\
Hybrid, regularised
&
$(2.474\pm0.083){\times}10^{-5}$
&
$0.0130\pm0.0004$
&
$3$
&
$0.0627\pm0.0012$
&
$0.0671\pm0.0034$
&
$0.1045\pm0.0031$
\\
Hybrid, unregularised
&
$(1.888\pm0.038){\times}10^{-5}$
&
$0.0114\pm0.0003$
&
$3$
&
$0.555\pm0.042$
&
$0.0879\pm0.0141$
&
$0.0778\pm0.0081$
\\
Symbolic, complete
&
$(1.573\pm0.017){\times}10^{-6}$
&
$(7.79\pm0.07){\times}10^{-4}$
&
$6$
&
$0$
&
$0.00462\pm0.00010$
&
$0.00344\pm0.00007$
\\
Neural only
&
$(1.746\pm0.216){\times}10^{-7}$
&
$(3.85\pm0.35){\times}10^{-4}$
&
--
&
--
&
$(1.70\pm0.48){\times}10^{-4}$
&
$0.00184\pm0.00012$
\\
\bottomrule
\end{tabular}
\end{table*}

\paragraph{Correction regularisation preserves the symbolic mechanism.}
Under the incomplete grammar, the regularised hybrid retains
\begin{equation*}
\label{eq:exp2-regularized-symbolic}
\dot q
=
(0.9581\pm0.0008)p,
\qquad
\dot p
=
-(0.7942\pm0.0024)\sin(q),
\end{equation*}
and has a normalised correction-energy ratio of $0.0627$. Without $\Rcorr$, the symbolic coefficients shrink to
\begin{equation*}
\label{eq:exp2-unregularized-symbolic}
\dot q
=
(0.3324\pm0.0065)p,
\qquad
\dot p
=
-(0.1718\pm0.0746)\sin(q),
\end{equation*}
while the normalised correction-energy ratio rises to $0.555$. Thus, an unrestricted correction largely takes over dynamics that the symbolic component could otherwise explain.

To evaluate whether the correction captures the mechanism omitted by the symbolic law, let
\begin{equation}
\tag{Eq.\ref{eq:complete_residual}}
r_p(q,p)
=
f_{\mathrm{true},p}(q,p)
-
F_{\mathcal E,\alpha,p}(q,p)
\end{equation}
be the complete residual left in the $p$ vector field. A scalar calibration is fitted using training data and then evaluated on held-out test transitions. Table~\ref{tab:exp2-allocation} shows that the regularised correction closely recovers both this total residual and the physical drag term. Figure~\ref{fig:exp2-main}(b) visualises the near one-to-one relationship between the calibrated correction and the complete symbolic residual. The drag-specific comparison is reported as a secondary diagnostic in Figure~\ref{fig:additional-diagnostics}(b). The unregularised model remains predictive, but its correction is much less specifically aligned with the omitted mechanism.

\begin{table*}[t]
\centering
\scriptsize
\setlength{\tabcolsep}{7pt}
\caption{Correction-allocation diagnostics in Experiment~2. Residual denotes the complete test-set $p$-field residual after the fitted symbolic law; drag denotes the isolated contribution $-0.4p|p|$. The reported $R^2$ values use a scalar calibration fitted on the training set.}
\label{tab:exp2-allocation}
\begin{tabular}{lccccc}
\toprule
Method
&
$\rho_{\mathrm{corr}}$
&
$\operatorname{Corr}(c_p,r_p)$
&
$R^2(c_p,r_p)$
&
$\operatorname{Corr}(c_p,\mathrm{drag})$
&
$R^2(c_p,\mathrm{drag})$
\\
\midrule
Hybrid, regularised
&
$0.0627\pm0.0012$
&
$0.9969\pm0.0002$
&
$0.9937\pm0.0004$
&
$0.9495\pm0.0014$
&
$0.9011\pm0.0028$
\\
Hybrid, unregularised
&
$0.555\pm0.042$
&
$0.9765\pm0.0023$
&
$0.9532\pm0.0047$
&
$0.4357\pm0.0561$
&
$0.1885\pm0.0506$
\\
\bottomrule
\end{tabular}
\end{table*}

The unregularised hybrid attains a numerically lower one-step state MSE,
$(1.888\pm0.038)\times10^{-5}$ compared with
$(2.474\pm0.083)\times10^{-5}$ for the regularised hybrid, and a lower mean
OOD rollout MSE, $0.0778\pm0.0081$ compared with $0.1045\pm0.0031$, as shown
in Table~\ref{tab:exp2-main}. Conversely, the regularised hybrid achieves the
lower test rollout MSE, $0.0671\pm0.0034$ compared with $0.0879\pm0.0141$.
Consequently, $\Rcorr$ should not be interpreted as providing an unconditional
accuracy improvement. Its role is to control symbolic--neural allocation: the
symbolic component retains the dominant reusable mechanism, while the
correction is reserved for residual dynamics that the grammar does not express
adequately.

\subsection{Experiments summary and scope}

The two experiments support complementary parts of the proposed framework.

Experiment~1 shows that operator complexity can guide predictive coordinate selection. Relative to symbolic regression fitted post hoc to frozen Neural JEPA coordinates, joint symbolic \ours reduces mean weighted symbolic complexity from $26.0$ to $4.67$, a factor of approximately $5.6$ (Table~\ref{tab:exp1-main}). It also reduces physical-state test rollout MSE from $1.229$ to $0.467$, OOD rollout MSE from $9.262$ to $3.435$, and OOD divergence rate from $0.178$ to $0.017$. Across all three seeds, the discovered laws retain the same dominant oscillator-like cross-coupling between $z_1$ and $z_2$. The within-representation one-step latent MSE is also approximately $3.6$ times lower, although this comparison is descriptive because latent MSE depends on the scale and coordinate system of each learned representation.

The collapse diagnostics establish the complementary necessity of representation constraints. Removing $\RIB$ produces nearly constant embeddings and a zero vector field, which implements an identity transition under the residual parameterisation. The constructive fixed-point control reaches the same degenerate solution directly. Extremely low latent prediction error and zero symbolic complexity can therefore indicate representation collapse rather than successful dynamics discovery, confirming the shortcut identified in Proposition~\ref{prop:collapse}.

These findings do not imply that \ours universally outperforms a neural predictor. Neural JEPA remains the most accurate model in raw one-step and rollout prediction. Moreover, the jointly learned symbolic coordinates are less affinely aligned with the original physical state than the prediction-only coordinates. Experiment~1 therefore demonstrates a trade-off: \ours sacrifices some predictive flexibility and simple physical-state alignment in exchange for a substantially more concise symbolic transition with lower recursive rollout error and divergence than the post-hoc symbolic baseline.

Experiment~2 separates grammar adequacy and symbolic--neural allocation from representation learning. When the grammar contains the required quadratic-drag primitive, the symbolic model selects the intended governing structure in every seed and recovers all coefficients to within approximately $7\%$ of their true values. Its rollout error approaches that of the neural predictor while retaining an explicit equation, showing that a sufficiently expressive grammar can remove the need for a residual correction in this controlled setting.

Under the incomplete grammar, the regularised hybrid retains the representable pendulum terms and has a normalised correction-energy ratio of $0.0627$, compared with $0.555$ for the unregularised hybrid. After scalar calibration on the training set, the regularised correction predicts the held-out complete symbolic residual with $R^2=0.994$ and is strongly aligned with the deliberately omitted quadratic-drag mechanism, with drag-specific $R^2=0.901$. Without $\Rcorr$, the symbolic coefficients shrink substantially and the correction absorbs much of the dynamics that the grammar can already represent.

Correction regularisation is nevertheless not an unconditional accuracy improvement. The unregularised hybrid attains a numerically lower one-step state MSE and lower mean OOD rollout MSE in this experiment, whereas the regularised hybrid attains the lower test rollout MSE. The role of $\Rcorr$ is therefore explanatory allocation: it encourages the symbolic component to retain compact representable structure and reserves the neural component for the remaining residual, rather than guaranteeing the smallest prediction error under every evaluation condition.

These experiments are controlled diagnostics rather than evidence of large-scale generality. They test coordinate selection, representation collapse, grammar adequacy, rollout behaviour, and symbolic--neural allocation in systems where the underlying state and omitted mechanism are known. Evaluation with visual observations at scale, genuinely pretrained foundation encoders, action-conditioned control tasks, real scientific datasets, and Bayesian posterior calibration remains future work.

\section{Discussion}

\paragraph{Key innovation.}
The contribution of \ours is not merely to insert symbolic regression into a JEPA predictor, and joint coordinate--equation learning has important precedents. Its distinctive contribution is to make the complexity of the induced transition operator an explicit criterion for learning a reconstruction-free predictive representation. The symbolic law is therefore not fitted only as a post-hoc explanation of fixed coordinates; in the end-to-end mode, it directly influences which predictive coordinates are selected. The constrained formulation, induced-dynamics complexity, and collapse analysis formalise this principle as learning the simplest adequate dynamics over an informative, non-collapsed state.

A second innovation is the controlled symbolic--neural decomposition
\(
\Hpred=\Fsym+c_\phi
\).
The symbolic component represents the dominant reusable mechanism, while the correction accounts for predictive structure that the selected grammar cannot express adequately. Because this decomposition is not identifiable from prediction loss alone, the symbolic-complexity and correction penalties encode an explicit allocation preference. The framework thereby addresses two coupled questions: which coordinates expose compact dynamics, and how predictable structure should be allocated between an explicit symbolic law and a flexible residual model.\footnote{At a broader level, this raises a more general question that we leave to separate work: given a task and dataset, which representation geometry or latent space is most suitable for JEPA learning? Different modelling objectives and selection criteria implicitly favour different latent geometries.}?

\paragraph{Empirical implications.}
The controlled experiments validate the two main deterministic consequences of the framework. First, allowing operator complexity to influence representation learning produces substantially more concise symbolic dynamics with lower rollout error and divergence than fitting equations post hoc, provided that explicit representation constraints prevent collapse. Second, correction regularisation controls how predictable structure is allocated between the symbolic law and neural residual under grammar misspecification. These findings establish a controllable trade-off among predictive fidelity, operator simplicity, representation quality, and symbolic--neural allocation rather than a universal improvement in raw predictive accuracy.

\paragraph{Interpretation of elegant dynamics.}
A compact symbolic transition can expose signs, interactions, symmetries, action couplings, and candidate invariants that are difficult to inspect in an unrestricted neural predictor. It can also be differentiated, simplified, checked against known constraints, and reused across trajectories. These advantages do not make a learned latent equation automatically physical, causal, or valid outside the observed regime. Its interpretation remains conditional on the learned coordinates, symbolic grammar, data support, predictive tolerance, and adequacy of the residual model.

The oscillator-like coordinates discovered in Experiment~1 illustrate both the value and the ambiguity of operator compression. The learned transition is concise and produces substantially lower rollout error than the post-hoc symbolic baseline, but the coordinates are nonlinearly related to the original physical state. Operator compression selects coordinates with simple evolution, not necessarily coordinates with direct physical semantics. Likewise, symbolic simplicity and linear simplicity are not mutually exclusive: the approximately linear oscillator discovered by \ours is a special case of compact symbolic dynamics. Comparisons with Koopman-style methods should therefore jointly evaluate predictive risk, operator complexity, coordinate quality, rollout behaviour, and downstream control utility under matched budgets.

These results motivate reporting prediction error, rollout stability, symbolic complexity, representation variation, coordinate alignment, and correction reliance together. No single metric is sufficient: low predictive error may accompany collapse, low symbolic complexity may reflect underfitting, and a small correction may simply indicate that an incomplete symbolic model has been left uncorrected.

\paragraph{Scope, modularity, and limitations.}
\ours is modular with respect to the encoder, representation regulariser, symbolic search procedure, correction architecture, and downstream planner. Experiment~1 evaluates the end-to-end mode in which representation and symbolic dynamics are learned jointly. Experiment~2 uses fixed, state-aligned coordinates and isolates the corresponding fixed-coordinate question: whether the supplied representation admits an adequate compact law and how unexplained structure is allocated to a correction. Direct experiments with genuinely pretrained JEPA or foundation encoders remain to be conducted.

The empirical validation is deliberately narrow. Both experiments use variants of one controlled pendulum system, two-dimensional states, prescribed symbolic libraries, and 3 optimisation seeds. They do not establish that compact symbolic latent laws will emerge in high-dimensional video, partially observed environments, scientific datasets, or embodied control tasks. The present differentiable sparse-library implementation also explores a more restricted model class than unrestricted expression-tree symbolic regression.

Simplicity is relative to the coordinate system, grammar, primitive weights, coefficient threshold, predictive tolerance, and observed regime. Alternating representation--operator optimisation is more expensive than fitting a single neural predictor, is susceptible to local optima, and does not guarantee recovery of a globally simplest adequate law. A correction with excessive capacity may conceal grammar misspecification, whereas an overly restricted correction may force residual effects into distorted symbolic coefficients. Moreover, discontinuous, stochastic, multiscale, or regime-switching systems may require local, piecewise, hierarchical, context-dependent, or distributional symbolic mechanisms rather than one compact global equation.

\paragraph{Bayesian and decision-making extensions.}
The present experiments validate only the deterministic framework. The Bayesian formulation provides uncertainty over symbolic structures, coefficients, residual covariance, and an optional Gaussian-process correction, but posterior calibration and decision-theoretic utility have not yet been evaluated. Similarly, the action-conditioned formulation exposes sensitivities, local linearisations, and an interface to external planning, but the paper reports no control experiment. Claims concerning posterior calibration, sample efficiency, control performance, or uncertainty-aware planning therefore remain hypotheses for future empirical study.

\section{Conclusion}

We introduced \ours, a reconstruction-free joint-embedding predictive framework that learns latent transitions as compact symbolic laws with optional neural corrections. Its central principle is to complement representation learning with operator compression: the representation must preserve an informative, non-collapsed predictive state, while the transition model should realise the simplest adequate law over that state. We formalised this principle through constrained operator compression and induced-dynamics complexity, characterised the coordinate non-identifiability of predictive representations, identified the collapse shortcut introduced by unconstrained dynamics simplification, and analysed how correction regularisation controls the allocation between symbolic and neural dynamics. The framework supports both alternating representation--equation learning and symbolic dynamics over fixed representations, with Bayesian and action-conditioned formulations providing extensions to uncertainty and decision making.

The controlled experiments support the two principal deterministic claims. Joint representation and symbolic-dynamics learning reduces symbolic complexity by approximately $5.6$ times and produces substantially lower physical-state rollout error and OOD divergence than symbolic regression fitted post hoc to prediction-only coordinates. The unconstrained one-step diagnostic yields near-constant embeddings and trivial identity dynamics, demonstrating that the predicted collapse shortcut is attainable and that operator simplicity is meaningful only for an admissible representation. Under controlled grammar misspecification, a complete grammar recovers the governing structure and coefficients consistently, while correction regularisation reduces the normalised correction-energy ratio from $0.555$ to $0.0627$ and yields a residual correction with calibrated test $R^2=0.994$, while limiting unnecessary reliance on the correction for representable dynamics.

These findings expose a controllable trade-off rather than a universal accuracy advantage. Flexible neural predictors remain more accurate, learned symbolic coordinates need not be simply aligned with physical variables, and correction regularisation may exchange some predictive flexibility for more controlled symbolic--neural allocation. The central conclusion is therefore that operator compression can select predictive coordinates whose induced dynamics are simple yet adequate, while representation constraints and correction control ensure that this simplicity is achieved by compressing the transition law rather than collapsing the representation or delegating the dynamics to the neural correction.

\bibliography{references}
\bibliographystyle{iclr2026_conference}

\appendix

\section{Objectives and Representation Constraints}
\label{app:objectives}

\subsection{Notation}

\begin{table}[h]
\centering
\small
\caption{Main notation. Operator-complexity quantities are defined relative to the selected latent normalisation, symbolic grammar, correction class, and evaluation distribution.}
\begin{tabularx}{\textwidth}{p{0.25\textwidth}X}
\toprule
Symbol & Meaning \\
\midrule
$X_C,X_T$
&
Context and target observations or temporal segments.
\\
$Z_C,Z_T$
&
Context and target embeddings in a common ambient latent space.
\\
$\epsinfo$
&
Predictor side information, such as masks, time offsets, actions, goals, or interventions.
\\
$E_\theta,E_{\bar\theta}$
&
Context and target encoders; $E_{\bar\theta}$ may be an EMA copy of $E_\theta$.
\\
$\mathcal E,\alpha$
&
Symbolic expression structure and its numerical coefficients.
\\
$\Fsym,c_\phi,\Hpred$
&
Symbolic component, neural correction, and complete direct-transition predictor, with
$\Hpred=\Fsym+c_\phi$.
\\
$V_{\mathcal E,\alpha,\phi},
\mathcal I_{\Delta t},
\Hpredt$
&
Hybrid latent vector field, integration operator, and resulting temporal transition,
$\Hpredt
=
\mathcal I_{\Delta t}
(z,V_{\mathcal E,\alpha,\phi},\epsinfo)$.
\\
$\Csym,\Rcorr,\Gamma$
&
Symbolic complexity, correction regulariser, and combined operator-complexity score,
$\Gamma=\Csym+\lambda_{\mathrm c}\Rcorr$.
\\
$\RIB$
&
Information-bottleneck or representation-quality regulariser.
\\
$\Theta_{\mathrm{repr}}$
&
Admissible class of informative, non-collapsed representations satisfying the selected latent-scale convention.
\\
$\Cdyn(E_\theta,E_{\bar\theta})$
&
Minimum adequate operator-complexity score induced by the encoders.
\\
$\Lpred,\Lroll$
&
One-step predictive risk and multi-step rollout risk.
\\
$\delta_{\mathrm{pred}},\delta_{\mathrm{roll}}$
&
Optional one-step and rollout adequacy tolerances; $\varepsilon$ is not used for tolerances.
\\
$\rho_{\mathrm{corr}}$
&
Normalised correction-energy ratio.
\\
\bottomrule
\end{tabularx}
\end{table}

\subsection{Temporal Objectives and Correction Diagnostics}

For temporal tuples
\begin{equation*}
\left(
X_{\le t},
X_{t+1:t+K},
\epsinfo_{t:t+K-1}
\right),
\end{equation*}
and, for continuous-time models, step sizes
$\Delta t_{t:t+K-1}$, define
\begin{align}
z_t
&=
E_\theta(X_{\le t}),
\\
z_{t+h}^{\mathrm{tar}}
&=
E_{\bar\theta}(X_{t+h}),
\qquad
h=1,\ldots,K.
\end{align}
The target encoder supplies slowly moving prediction targets in the same ambient latent space as the context encoder.

To cover both temporal realisations, define the complete one-step transition used at rollout step $h$ by
\begin{equation}
\label{eq:complete-temporal-transition}
\mathcal H_{t+h}(z)
=
\begin{cases}
\Hpred
\left(
z,
\epsinfo_{t+h}
\right),
&
\text{direct-transition realisation},
\\[4pt]
\mathcal I_{\Delta t_{t+h}}
\left(
z,
V_{\mathcal E,\alpha,\phi},
\epsinfo_{t+h}
\right),
&
\text{vector-field realisation}.
\end{cases}
\end{equation}
The free recursive rollout is
\begin{equation}
\label{eq:general-free-rollout}
\widehat z_t
=
z_t,
\qquad
\widehat z_{t+h+1}
=
\mathcal H_{t+h}
\left(
\widehat z_{t+h}
\right),
\qquad
h=0,\ldots,K-1.
\end{equation}

For the forward-Euler realisation used in the reported experiments,
Equation~\eqref{eq:complete-temporal-transition} becomes
\begin{equation}
\label{eq:appendix-euler-transition}
\mathcal H_{t+h}(z)
=
z
+
\Delta t_{t+h}
\left[
F_{\mathcal E,\alpha}
\left(
z,
\epsinfo_{t+h}
\right)
+
c_\phi
\left(
z,
\epsinfo_{t+h}
\right)
\right].
\end{equation}

The rollout loss is
\begin{equation}
\label{eq:rollout-loss}
\Lroll
=
\sum_{h=1}^{K}
w_h
d
\left(
\widehat z_{t+h},
\sg
\left(
z_{t+h}^{\mathrm{tar}}
\right)
\right),
\qquad
w_h\ge0.
\end{equation}
Teacher-forced and free-rollout terms may be combined. For temporal applications in which recursive accuracy is part of the required notion of adequacy, the constrained formulation may additionally impose
\begin{equation*}
\Lroll
\le
\delta_{\mathrm{roll}}.
\end{equation*}
The task-general definition retains only the one-step constraint because recursive rollout is not defined for every JEPA context--target relation.

When correction control is applied during temporal rollout, the correction output may be penalised at every recursively visited state. In the vector-field realisation, the correction penalty and diagnostics are evaluated on the raw vector-field correction $c_\phi$ before numerical integration and before multiplication by $\Delta t$. This prevents them from depending artificially on the integration step size.

The correction-energy ratio is evaluated on the decomposed predictive object itself: the complete predicted target in the direct-transition realisation, or the latent vector field in the continuous-time realisation. Define
\begin{equation}
\label{eq:decomposed-predictive-object}
\mathcal G_{\mathcal E,\alpha,\phi}(z,\epsinfo)
=
\begin{cases}
\Hpred(z,\epsinfo),
&
\text{direct-transition realisation},
\\[3pt]
V_{\mathcal E,\alpha,\phi}(z,\epsinfo),
&
\text{vector-field realisation}.
\end{cases}
\end{equation}
The normalised correction-energy ratio is
\begin{equation}
\label{eq:correction-energy-ratio-appendix}
\rho_{\mathrm{corr}}
=
\frac{
\E
\left[
\left\|
c_\phi(Z_C,\epsinfo)
\right\|_2^2
\right]
}{
\E
\left[
\left\|
\mathcal G_{\mathcal E,\alpha,\phi}
(Z_C,\epsinfo)
\right\|_2^2
\right]
+
\eta_0
},
\qquad
\eta_0>0.
\end{equation}
For the vector-field experiments, this reduces to
\begin{equation}
\label{eq:correction-energy-ratio-appendix_vec}
\rho_{\mathrm{corr}}
=
\frac{
\E
\left[
\left\|
c_\phi(z)
\right\|_2^2
\right]
}{
\E
\left[
\left\|
F_{\mathcal E,\alpha}(z)
+
c_\phi(z)
\right\|_2^2
\right]
+
\eta_0
}.
\end{equation}
The residual identity contribution $z$ in Equation~\eqref{eq:appendix-euler-transition} and the factor $\Delta t$ are excluded because they are not part of the symbolic--neural decomposition of the vector field.

A low value of $\rho_{\mathrm{corr}}$ indicates limited reliance on the correction, but is not automatically preferable: a near-zero correction paired with high predictive error may indicate underfitting. The ratio is therefore interpreted jointly with predictive risk and symbolic complexity. Because the symbolic and correction components need not be orthogonal, $\rho_{\mathrm{corr}}$ is a normalised energy ratio rather than an additive fraction of the total dynamics.

\subsection{Representation Regularisation}
\label{app:ibforms}

The theory requires an admissible, non-collapsed predictive representation. Different implementations impose different assumptions.

\begin{table}[h]
\centering
\small
\caption{Representative implementations of $\RIB$. VICReg and Barlow Twins are practical surrogates for representation quality and non-collapse, not exact estimators of the mutual-information bottleneck.}
\begin{tabularx}{\textwidth}{p{0.18\textwidth}p{0.30\textwidth}X}
\toprule
Form & Example objective & Role in \ours \\
\midrule
Predictive mutual information
&
$\I(X_C;Z_C)-\beta_{\mathrm{rel}}\I(Z_C;Z_T)$
&
Direct statement of relevant compression; difficult to estimate in high dimension.
\\
Variational bottleneck
&
$\E_x\KL(q_\theta(z\mid x)\|p_0(z))$ plus predictive loss
&
Controls representational rate and supports stochastic encoders.
\\
VICReg-style
&
invariance $+$ variance $+$ covariance
&
Prevents constant collapse, encourages consistency under selected augmentations, and reduces linear redundancy.
\\
Barlow Twins-style
&
cross-correlation toward identity
&
Encourages invariance and redundancy reduction without an explicit rate term.
\\
Dimensional bottleneck
&
fixed low $d_z$ or rank penalty
&
Simple capacity control; may remove relevant variables if too restrictive.
\\
Latent noise
&
$\widetilde z=z+\eta$, $\eta\sim p_\eta$
&
Discourages fragile encodings and can regularise symbolic search.
\\
Explicit admissibility
&
$\Var(Z_j)\ge v_{\min}$ and off-diagonal covariance bounds
&
Directly rules out constant and highly redundant representations.
\\
\bottomrule
\end{tabularx}
\end{table}

A concrete VICReg-style instance is
\begin{equation}
\label{eq:vicreg}
\RIB^{\mathrm{VIC}}
=
\lambda_{\mathrm{inv}}
\mathcal L_{\mathrm{inv}}
+
\lambda_{\mathrm{var}}
\mathcal L_{\mathrm{var}}
+
\lambda_{\mathrm{cov}}
\mathcal L_{\mathrm{cov}}.
\end{equation}
The three terms are the usual invariance, per-coordinate variance-floor, and off-diagonal covariance penalties. In \ours, the variance term blocks the constant-coordinate shortcut of Proposition~\ref{prop:collapse}; the covariance term discourages redundant linear dependence between coordinates; and the invariance term encourages consistency under the specified observation augmentations. The experimental instantiation in Appendix~\ref{app:experimental-details} additionally includes a weak mean-centering term
$\lambda_{\mathrm{mean}}\mathcal L_{\mathrm{mean}}$.

\section{Proofs}
\label{app:proofs}

\subsection{Predictive Coordinates and Collapse}

\subsubsection{Proof of Proposition~\ref{prop:nonident}}
\label{app:proof-nonident}

For almost every training tuple,
\begin{align*}
\widetilde F(\widetilde E(X_C),\epsinfo)
&=
h
\left(
F
\left(
h^{-1}(h(E(X_C))),
\epsinfo
\right)
\right)
\\
&=
h(F(E(X_C),\epsinfo))
\\
&=
h(E(X_T))
\\
&=
\widetilde E(X_T).
\end{align*}
Thus, the transformed representation is equally exact. Symbolic complexity is not invariant under general conjugacies: a simple linear or polynomial $F$ can become a complicated rational, transcendental, or non-representable $\widetilde F$, and conversely. Therefore, predictive exactness alone does not identify coordinates, while an operator-complexity criterion can distinguish them. \hfill$\square$

\subsubsection{Proof of Proposition~\ref{prop:collapse}}
\label{app:proof-collapse}

Under the constant encoders,
\begin{equation*}
Z_C
=
Z_T
=
z_0
\end{equation*}
for every example. By assumption,
\begin{equation*}
H_0(z_0,\epsinfo)
=
z_0
\end{equation*}
for every admissible $\epsinfo$. Hence
\begin{equation*}
d
\left(
H_0(Z_C,\epsinfo),
Z_T
\right)
=
d(z_0,z_0)
=
0,
\end{equation*}
and therefore $\Lpred=0$. Every feasible model has nonnegative predictive loss, while $H_0$ attains the minimum weighted operator penalty by assumption. Consequently, $(E_0,E_0,H_0)$ attains the global lower bound of the unconstrained objective. \hfill$\square$

\subsection{Symbolic--Neural Allocation}

\subsubsection{Proof of Proposition~\ref{prop:decomp}}
\label{app:proof-decomp}

Direct substitution gives
\begin{equation*}
(\Fsym+r)+(c_\phi-r)
=
\Fsym+c_\phi.
\end{equation*}
Both decompositions therefore induce identical predictions on every input and have identical predictive risk. Unless the penalties or function classes distinguish them, the allocation of structure is not identifiable. \hfill$\square$

\subsubsection{Proof of Proposition~\ref{prop:correction}}
\label{app:proof-correction}

Condition on $(Z_C,\epsinfo)=(z,e)$ and write
$m=m(z,e)$ and $a=F(z,e)$. For any correction value $c$,
\begin{align}
\E[\|Z_T-a-c\|_2^2\mid z,e]
&=
\E[\|Z_T-m\|_2^2\mid z,e]
+
\|m-a-c\|_2^2,
\end{align}
because the conditional cross term vanishes. The first term does not depend on $c$. Hence the conditional objective is
\begin{equation*}
q(c)
=
\|m-a-c\|_2^2
+
\lambda\|c\|_2^2.
\end{equation*}
It is strictly convex for $\lambda>0$, and
\begin{equation*}
\nabla_c q(c)
=
-2(m-a-c)
+
2\lambda c.
\end{equation*}
Setting the gradient to zero gives
$(1+\lambda)c=m-a$, proving Equation~\eqref{eq:correction-shrinkage}. Integrating the pointwise minimiser over $(Z_C,\epsinfo)$ completes the proof. \hfill$\square$

\subsection{Rollout and Bayesian Results}

\subsubsection{Proof of Proposition~\ref{prop:rollout}}
\label{app:proof-rollout}

For the same action $u_{t+h}$ in the true and learned rollouts,
\begin{align*}
e_{h+1}
&=
\|T(z_{t+h},u_{t+h})-H(\widehat z_{t+h},u_{t+h})\|_2
\\
&\le
\|T(z_{t+h},u_{t+h})-H(z_{t+h},u_{t+h})\|_2
\\
&\quad+
\|H(z_{t+h},u_{t+h})-H(\widehat z_{t+h},u_{t+h})\|_2
\\
&\le
\delta+Le_h.
\end{align*}
Iterating this recursion yields
\begin{equation*}
e_h
\le
L^he_0
+
\delta
\sum_{j=0}^{h-1}L^j.
\end{equation*}
Evaluating the geometric sum gives Equation~\eqref{eq:rollout-bound}. \hfill$\square$

\subsubsection{Proof of Theorem~\ref{thm:map}}
\label{app:proof-map}

Under the conditionally independent Gaussian model in Equation~\eqref{eq:bayes-like} and the isotropic covariance assumption $\Sigma=\sigma^2 I$, the negative log-likelihood is
\begin{align*}
-\log p(\D_Z\mid\mathcal E,\alpha,\sigma^2 I)
&=
\frac{nd_z}{2}\log(2\pi\sigma^2)
\\
&\quad+
\frac{1}{2\sigma^2}
\sum_{i=1}^n
\left\|
z_T^{(i)}
-
F_{\mathcal E,\alpha}
\left(
z_C^{(i)},
\epsinfo^{(i)}
\right)
\right\|_2^2,
\end{align*}
where $d_z$ is the target-embedding dimension.

From the structure prior in Equation~\eqref{eq:structure-prior},
\begin{equation*}
-\log p(\mathcal E)
=
\gamma\Csym(\mathcal E)
+
\mathrm{const}.
\end{equation*}
Bayes' rule gives
\begin{align*}
-\log p(\mathcal E,\alpha\mid\D_Z)
&=
-\log p(\D_Z\mid\mathcal E,\alpha,\sigma^2 I)
-\log p(\alpha\mid\mathcal E)
-\log p(\mathcal E)
+
\mathrm{const}
\\
&=
\frac{1}{2\sigma^2}
\sum_{i=1}^n
\left\|
z_T^{(i)}
-
F_{\mathcal E,\alpha}
\left(
z_C^{(i)},
\epsinfo^{(i)}
\right)
\right\|_2^2
\\
&\quad+
\gamma\Csym(\mathcal E)
-
\log p(\alpha\mid\mathcal E)
+
\mathrm{const},
\end{align*}
where the final constant is independent of $(\mathcal E,\alpha)$. Therefore, maximising the posterior is equivalent to minimising the objective in Equation~\eqref{eq:map-objective}. \hfill$\square$

\section{Symbolic Model and Optimisation}
\label{app:optimization}

\subsection{Grammar and Structural Complexity}
\label{app:grammar}

This subsection describes a general expression-tree realisation of \ours. The reported experiments instead use the fixed differentiable symbolic libraries specified in Appendix~\ref{app:experimental-details}.

A grammar should be expressive enough to represent plausible mechanisms but small enough to make search and structural analysis meaningful. A generic typed grammar is
\begin{align*}
\mathcal E
::={}&
\alpha
\mid
z_j
\mid
\epsinfo_k
\mid
\mathcal E+\mathcal E
\mid
\mathcal E-\mathcal E
\mid
\mathcal E\times\mathcal E
\\
&\mid
\operatorname{safeDiv}(\mathcal E,\mathcal E)
\mid
\sin(\mathcal E)
\mid
\cos(\mathcal E)
\mid
\exp_{\mathrm{clip}}(\mathcal E)
\mid
\log_{\mathrm{safe}}(\mathcal E).
\end{align*}
Domain-safe operators avoid undefined evaluations during search. For vector dynamics, one expression is learned per target coordinate, with optional shared subexpressions.

A weighted-tree complexity may satisfy
\begin{equation*}
w_{\mathrm{constant}}
<
w_{\mathrm{linear}}
<
w_{\mathrm{product}}
<
w_{\mathrm{division}}
<
w_{\mathrm{transcendental}},
\end{equation*}
or use a code length derived from operator frequencies. Complexity weights must be fixed before test evaluation or tuned on validation data; otherwise, equation simplicity can be selected post hoc.

For unrestricted expression-tree search, equivalent expressions should be canonicalised before scoring. Recommended steps include constant folding, commutative sorting, algebraic simplification, removal of neutral elements, coefficient thresholding, and numerical equivalence checks on held-out points. In the reported sparse-library experiments, the feature ordering is fixed, and canonicalisation reduces primarily to coefficient thresholding and removal of inactive output--term pairs.

\subsection{Differentiable Sparse-Library Realisation}

For discrete expression-tree search,
$\Csym^{\mathrm{train}}$
may equal the structural complexity
$\Csym(\mathcal E)$
in Equation~\eqref{eq:omega}. For the differentiable sparse-library implementation used in the experiments, we use
\begin{equation}
\label{eq:smooth-symbolic-complexity}
\Csym^{\mathrm{train}}(\mathcal E,\alpha)
=
\sum_{j=1}^{d_z}
\sum_{k=1}^{P}
w_k
\sqrt{\alpha_{kj}^{\,2}+\xi_\alpha},
\qquad
\xi_\alpha>0,
\end{equation}
where $\alpha_{kj}$ is the coefficient of library term $k$ in output coordinate $j$, and $P$ denotes the number of candidate library terms. This smooth quantity guides optimisation, while the thresholded support is used to compute the reported structural complexity
$\Csym(\widehat{\mathcal E})$.

\subsection{Stage-Specific Objectives}
\label{app:stage-objectives}

The practical training objective in Equation~\eqref{eq:lagrangian} contains five common components. Each optimisation stage restricts its trainable variables and omits terms that are constant or inapplicable.

\paragraph{Representation warm start.}
Before symbolic search, the encoder and a neural predictor are trained using
\begin{equation}
\label{eq:warm-objective}
\mathcal L_{\mathrm{warm}}
=
\Lpred
+
\lambda_{\mathrm r}\Lroll
+
\lambda_{\mathrm{IB}}\RIB.
\end{equation}
The target encoder is updated by exponential moving average.

\paragraph{Dynamics search.}
With the encoders fixed, the dynamics phase minimises
\begin{equation}
\label{eq:dynamics-search-objective}
\mathcal L_{\mathrm{dyn}}
=
\Lpred
+
\lambda_{\mathrm r}\Lroll
+
\lambda_{\mathrm s}
\Csym^{\mathrm{train}}(\mathcal E,\alpha)
+
\lambda_{\mathrm c}\Rcorr(\phi)
\end{equation}
over $(\mathcal E,\alpha,\phi)$.

\paragraph{Space search.}
With the discrete symbolic structure fixed or locally relaxed, the space phase minimises
\begin{equation}
\label{eq:space-search-objective}
\mathcal L_{\mathrm{space}}
=
\Lpred
+
\lambda_{\mathrm r}\Lroll
+
\lambda_{\mathrm{IB}}\RIB
+
\lambda_{\mathrm s}
\Csym^{\mathrm{train}}(\mathcal E,\alpha)
+
\lambda_{\mathrm c}\Rcorr(\phi)
\end{equation}
over the encoder and continuously optimised dynamics parameters. When both $\mathcal E$ and $\alpha$ are fixed, the symbolic-complexity term is constant and may be omitted. When $\alpha$ or a relaxed support remains trainable, the term remains active.

\paragraph{Frozen-encoder search.}
For a permanently frozen encoder, the objective becomes
\begin{equation}
\label{eq:frozen-search-objective}
\mathcal L_{\mathrm{frozen}}
=
\Lpred
+
\lambda_{\mathrm r}\Lroll
+
\lambda_{\mathrm s}
\Csym^{\mathrm{train}}(\mathcal E,\alpha)
+
\lambda_{\mathrm c}\Rcorr(\phi),
\end{equation}
optimised only over $(\mathcal E,\alpha,\phi)$.

\subsection{Relation Between Constrained and Penalised Formulations}

For completeness, consider a finite-dimensional convex relaxation with objective $\Gamma(w)$ and convex constraints $g_j(w)\le0$. If Slater's condition holds, strong duality gives multipliers $\lambda_j^\star\ge0$ such that every primal optimum minimises
\begin{equation*}
\Gamma(w)
+
\sum_j
\lambda_j^\star g_j(w).
\end{equation*}
This standard result motivates Equation~\eqref{eq:lagrangian}. It does not establish global equivalence for neural encoders, discrete expression search, or nonconvex correction models. In those settings, constraint-aware search and Pareto reporting are safer than interpreting one penalty weight as canonical.

Equation~\eqref{eq:lagrangian} should therefore be interpreted as a scalarisation of the constrained principle. The regularisation weights control trade-offs rather than recovering a unique constrained optimum. Models should be assessed jointly using predictive error, rollout error, symbolic complexity, representation-collapse diagnostics, and the normalised correction-energy ratio. Across multiple regularisation settings, this gives the empirical trade-off
\begin{equation*}
\label{eq:pareto-surface}
\left(
\Lpred,
\Lroll,
\Csym(\widehat{\mathcal E}),
\rho_{\mathrm{corr}}
\right),
\end{equation*}
subject to an admissible non-collapsed representation.

\section{Detailed Experimental Setup and Additional Results}
\label{app:experimental-details}

This section gives the complete implementation details for the two experiments in Section~\ref{sec:experiments}. Both experiments use the same continuous-time simulator, residual-predictor convention, trajectory counts, temporal windows, and validation-based checkpoint procedure unless stated otherwise.

\subsection{Experimental Protocol}

\paragraph{Simulation and data splits.}
The physical state is $s=(q,p)$, with vector field
\begin{equation}
f_\kappa(q,p)
=
\begin{bmatrix}
p\\
-\sin(q)-\kappa p|p|
\end{bmatrix}.
\end{equation}
Experiment~1 uses $\kappa=0$, whereas Experiment~2 uses $\kappa=0.4$. Trajectories are generated using fourth-order Runge--Kutta integration. At every transition, the time step is sampled independently from
\begin{equation*}
\Delta t
\in
\{0.025,0.04,0.06\}.
\end{equation*}

Each experiment uses $300$ training trajectories, $80$ validation trajectories, $80$ test trajectories, and $80$ OOD trajectories, with $100$ transitions per trajectory. The initial-state ranges are listed in Table~\ref{tab:experimental-ranges}.

\begin{table}[h]
\centering
\small
\caption{Initial-state ranges used to generate trajectories. Validation and test sets use independent trajectories from the same ranges.}
\label{tab:experimental-ranges}
\begin{tabular}{lccc}
\toprule
Experiment & Split & $q_0$ range & $p_0$ range \\
\midrule
\multirow{3}{*}{Experiment 1}
& Train & $[-1.8,1.8]$ & $[-1.4,1.4]$ \\
& Validation/test & $[-1.8,1.8]$ & $[-1.4,1.4]$ \\
& OOD & $[-2.65,2.65]$ & $[-2.0,2.0]$ \\
\midrule
\multirow{3}{*}{Experiment 2}
& Train & $[-2.2,2.2]$ & $[-2.4,2.4]$ \\
& Validation/test & $[-2.2,2.2]$ & $[-2.4,2.4]$ \\
& OOD & $[-2.9,2.9]$ & $[-3.0,3.0]$ \\
\bottomrule
\end{tabular}
\end{table}

The fixed dataset seeds are $1001$, $1004$, $1002$, and $1003$ for the Experiment~1 training, validation, test, and OOD sets, respectively. Experiment~2 uses $2001$, $2004$, $2002$, and $2003$. Model optimisation is repeated with seeds
\begin{equation*}
\{7,19,37\}.
\end{equation*}

\paragraph{Experiment~1 observation map.}
Experiment~1 does not provide the physical state as named input coordinates. Define the raw feature vector
\begin{align*}
\psi(q,p)
=
\big[
&
q,p,
\sin q,\cos q,\sin(2q),\cos(2q),
\sin p,\cos p,
\tanh q,\tanh p,
\\
&
qp,q^2,p^2,q^3/6,p^3/6,1,
\tanh(Ws+b),
\sin(Ws+b)
\big],
\end{align*}
where $W\in\mathbb R^{8\times2}$ and $b\in\mathbb R^8$ are fixed random parameters. The final observation is
\begin{equation}
X
=
Q\psi(q,p)
+
\xi,
\qquad
\xi
\sim
\mathcal N(0,0.003^2I),
\end{equation}
where $Q\in\mathbb R^{32\times32}$ is a fixed orthogonal matrix. The random observation map uses seed $101$. Every observation dimension is standardised using the mean and standard deviation of the training set only.

Because the raw coordinates $q$ and $p$ are included in $\psi(q,p)$ and $Q$ is orthogonal, the noiseless physical state is linearly recoverable from the complete observation. The nonlinear terms act as mixed distractor and auxiliary features. This controlled design is intended to study predictive-coordinate and operator selection rather than nonlinear observability.

\subsection{Model and Training Configuration}

\paragraph{Architectures.}
The Experiment~1 encoder is
\begin{equation*}
32
\longrightarrow
96
\longrightarrow
96
\longrightarrow
2,
\end{equation*}
with SiLU activations in the hidden layers. The target encoder is an exponential-moving-average copy with decay $0.99$.

The neural vector field has two hidden layers of width $96$ with SiLU activations and bounded output
\begin{equation*}
f_\psi(z)
=
3\tanh
\left(
\operatorname{MLP}_\psi(z)
\right).
\end{equation*}
The correction network in Experiment~2 has two hidden layers of width $48$, uses $\tanh$ activations, and has bounded output
\begin{equation*}
c_\phi(z)
=
\tanh
\left(
\operatorname{MLP}_\phi(z)
\right).
\end{equation*}

All predictors use the residual transition in Equation~\eqref{eq:experimental-residual-predictor}. Consequently, the zero vector field implements an identity transition:
\begin{equation*}
\widehat z_{t+1}
=
z_t.
\end{equation*}
If both encoders map every input to the same arbitrary constant $z_0$, this identity transition predicts every target embedding perfectly. The constant need not be the zero vector.

\paragraph{Representation regularisation.}
The practical representation term is
\begin{equation*}
\RIB^{\mathrm{exp}}
=
2\mathcal L_{\mathrm{inv}}
+
3\mathcal L_{\mathrm{var}}
+
0.25\mathcal L_{\mathrm{cov}}
+
0.01\mathcal L_{\mathrm{mean}}.
\end{equation*}
The invariance term compares two observation-noise augmentations with standard deviation $0.012$. The variance term imposes a unit-scale coordinate-variance target, the covariance term penalises off-diagonal covariance, and the mean term weakly centres the representation. The total representation term has unit weight in the practical objective.

\paragraph{Symbolic libraries and coefficient initialisation.}
For Experiment~1, the symbolic feature library is
\begin{equation*}
\Theta(z)
=
\left[
1,z_1,z_2,
\sin z_1,\sin z_2,
\cos z_1,\cos z_2,
z_1^2,z_2^2,z_1z_2
\right].
\end{equation*}
The corresponding complexity weights are
\begin{equation*}
(0.5,1,1,2,2,2,2,1.5,1.5,2).
\end{equation*}

Experiment~2 uses an output-specific mask rather than permitting every feature in every equation. The incomplete grammar is
\begin{equation*}
\dot q:\{p\},
\qquad
\dot p:\{\sin(q)\},
\end{equation*}
whereas the complete grammar is
\begin{equation*}
\dot q:\{p\},
\qquad
\dot p:\{\sin(q),p|p|\}.
\end{equation*}
The complexity weight of $p|p|$ is $3$. Linear $p$ is not permitted in the equation for $\dot p$.

Coefficients are initialised by sequentially thresholded ridge regression. The ridge parameter is $10^{-5}$ and the STLSQ threshold is $0.035$. Continuous training uses
\begin{equation*}
\Csym^{\mathrm{train}}
=
\sum_{j,k}
w_k
\sqrt{\alpha_{kj}^2+10^{-8}}.
\end{equation*}
After each symbolic phase, coefficients with magnitude below $0.0175$ are set to zero. Final equations and discrete complexity are reported using the stricter threshold
\begin{equation*}
|\alpha_{kj}|
\ge
0.05.
\end{equation*}

\paragraph{Optimisation settings.}

\begin{table*}[t]
\centering
\small
\caption{Optimisation settings used in the full three-seed experiments.}
\label{tab:experimental-hyperparameters}
\begin{tabularx}{\textwidth}{p{0.27\textwidth}p{0.18\textwidth}X}
\toprule
Parameter & Value & Use \\
\midrule
Batch size
&
$256$
&
All training phases
\\
Training rollout horizon
&
$10$
&
Free recursive rollout loss
\\
Rollout weight
&
$0.35$
&
Warm JEPA, Neural JEPA, post-hoc symbolic, joint \ours, and all Experiment~2 models
\\
Collapse-diagnostic rollout weight
&
$0$
&
One-step collapse and fixed-point diagnostic conditions
\\
Evaluation horizon
&
$80$
&
Open-loop evaluation on $60$ trajectories
\\
Encoder learning rate
&
$5{\times}10^{-4}$
&
Warm representation training
\\
Neural continuation scale
&
$0.5$
&
Effective encoder rate $2.5{\times}10^{-4}$
\\
\ours space-search scale
&
$0.35$
&
Effective encoder rate $1.75{\times}10^{-4}$
\\
One-step collapse-diagnostic encoder scale
&
$0.35$
&
Collapse diagnostic
\\
Fixed-point encoder scale
&
$0.20$
&
Constructive constant-representation control
\\
Neural/correction learning rate
&
$8{\times}10^{-4}$
&
Neural vector fields and corrections
\\
Symbolic coefficient learning rate
&
$2{\times}10^{-3}$
&
Continuous coefficient fitting
\\
AdamW weight decay
&
$10^{-5}$
&
All optimisers
\\
Gradient clipping
&
$3.0$
&
Global gradient norm
\\
Experiment~1 $\lambda_{\mathrm s}$
&
$2.5{\times}10^{-4}$
&
Symbolic complexity
\\
Default joint $\lambda_{\mathrm c}$
&
$2.0{\times}10^{-2}$
&
Available in the general implementation but inactive for symbolic-only Experiment~1
\\
Experiment~2 $\lambda_{\mathrm s}$
&
$1.5{\times}10^{-4}$
&
Symbolic complexity
\\
Experiment~2 regularised $\lambda_{\mathrm c}$
&
$1.5{\times}10^{-2}$
&
Hybrid correction control
\\
Correction parameter penalty
&
$10^{-6}\|\phi\|_2^2$
&
Capacity component of $\Rcorr$
\\
\bottomrule
\end{tabularx}
\end{table*}

The Experiment~1 warm neural JEPA is trained for at most $130$ epochs. Neural JEPA is then continued for at most $90$ epochs. Post-hoc symbolic regression trains the symbolic dynamics for at most $65$ epochs with both encoders frozen.

Joint symbolic \ours performs four alternating cycles. Each cycle consists of at most $65$ epochs of dynamics search with frozen encoders and at most $85$ epochs of representation-space search. The one-step collapse diagnostic is trained for at most $100$ epochs using
\begin{equation*}
\mathcal L_{\mathrm{collapse}}
=
\Lpred
+
\lambda_{\mathrm s}
\Csym^{\mathrm{train}},
\end{equation*}
with neither $\RIB$ nor $\Lroll$. The constructive fixed-point control is initialised at
\begin{equation*}
z_0
=
(0.65,-0.35)
\end{equation*}
and trained for at most five epochs with the same one-step objective. Each Experiment~2 model is trained for at most $260$ epochs.

\subsection{Validation and Evaluation}

\paragraph{Checkpoint and cycle selection.}
Validation is performed every five epochs using at most sixteen validation batches. Define
\begin{equation*}
\mathcal R_{\mathrm{val}}
=
\Lpred^{\mathrm{val}}
+
0.35\Lroll^{\mathrm{val}}.
\end{equation*}
For Experiment~1 conditions that update the encoder, checkpoints failing the prescribed non-collapse thresholds are ineligible whenever at least one eligible checkpoint is available. The one-step collapse and fixed-point diagnostic conditions are exempt from this rule. Eligible checkpoints are selected lexicographically:

\begin{enumerate}[leftmargin=*,itemsep=1pt]
\item lower validation predictive risk is preferred;
\item when two risks are within a relative tolerance of $2\%$, lower discrete symbolic complexity is preferred;
\item the regularised validation objective and then the earlier epoch break any remaining tie.
\end{enumerate}

For checkpoint selection, a representation is considered non-collapsed when
\begin{equation*}
\min_j\operatorname{Std}(Z_j)
\ge
0.20
\qquad\text{and}\qquad
\operatorname{Tr}(\operatorname{Cov}Z)
\ge
0.10.
\end{equation*}
The one-step collapse and explicit fixed-point controls are selected without imposing this constraint because their purpose is to measure collapse. Their observed collapse statistics are nevertheless recorded.

The same validation rule is applied within every standard training phase, and the best completed alternating cycle is restored for final testing. The selected Experiment~1 cycles are
\begin{equation*}
c^\star_{7}
=
3,
\qquad
c^\star_{19}
=
1,
\qquad
c^\star_{37}
=
4.
\end{equation*}

\paragraph{Evaluation metrics.}
Experiment~1 reports:
\begin{itemize}[leftmargin=*,itemsep=1pt]
\item latent one-step MSE against the corresponding next-state target embedding; end-to-end JEPA models use the EMA target embedding, whereas the post-hoc symbolic baseline uses next-state embeddings from the same frozen context encoder used to construct its transition dataset;
\item affine-aligned physical-state rollout MSE, using an affine map fitted only on training latents;
\item minimum latent coordinate standard deviation and covariance trace;
\item affine state-probe $R^2$;
\item thresholded symbolic complexity;
\item clipped rollout MSE and divergence rate.
\end{itemize}

A rollout is marked divergent when the latent norm exceeds $40$, the mapped state norm exceeds $25$, or a non-finite value occurs. Diverged or missing errors are assigned the clipping value $100$ when computing the clipped mean.

Experiment~2 reports the same state and rollout metrics directly in $(q,p)$ coordinates. It uses the normalised correction-energy ratio defined in Equation~\eqref{eq:experimental-correction-share}, with expectations computed over independent test states. The residual identity contribution $z_t$ is excluded from the denominator because including it would make the diagnostic depend on the arbitrary scale of the state coordinates.

For the $p$ coordinate, define the symbolic residual
\begin{equation*}
r_p(z)
=
f_{\mathrm{true},p}(z)
-
F_{\mathcal E,\alpha,p}(z).
\end{equation*}
A scale-only calibration
\begin{equation*}
a^\star
=
\argmin_a
\E_{\mathrm{train}}
\left[
\left(
r_p(z)
-
a c_{\phi,p}(z)
\right)^2
\right]
\end{equation*}
is fitted on training transitions. Correlation and calibrated $R^2$ are then evaluated on the independent test set. The same procedure is applied to the isolated drag target
\begin{equation*}
-0.4p|p|.
\end{equation*}

\subsection{Post-Hoc Symbolic Baseline}
\label{app:posthoc-symbolic}

The post-hoc symbolic baseline separates representation learning from equation discovery. It tests whether coordinates learned solely for neural prediction already admit a compact symbolic transition, without allowing symbolic simplicity to influence the representation itself.

\paragraph{Stage 1: learning prediction-oriented coordinates.}
We first train Neural JEPA by jointly optimising the context encoder and neural residual vector field. Given an observation $x_t$, the selected model produces
\begin{equation}
z_t
=
E_{\theta^\star}(x_t),
\end{equation}
and predicts the next latent state according to
\begin{equation}
\widehat z_{t+1}
=
z_t
+
\Delta t_t f_{\psi^\star}(z_t).
\end{equation}
The Neural JEPA checkpoint is selected using the held-out validation risk defined in Equation~\eqref{eq:validation-risk}. At this stage, the representation is optimised for predictive accuracy and representation quality, but not for the complexity of a subsequently fitted symbolic equation.

\paragraph{Stage 2: freezing the learned coordinates.}
After model selection, the encoder parameters $\theta^\star$ are frozen. The training trajectories are encoded to construct
\begin{equation}
\mathcal D_Z
=
\left\{
\left(
z_t^{(i)},
z_{t+1}^{(i)},
\Delta t_t^{(i)}
\right)
\right\}_{i=1}^{N},
\end{equation}
where
\begin{equation}
z_t^{(i)}
=
E_{\theta^\star}
\left(
x_t^{(i)}
\right),
\qquad
z_{t+1}^{(i)}
=
E_{\theta^\star}
\left(
x_{t+1}^{(i)}
\right).
\end{equation}
Using the same frozen context encoder at both endpoints ensures that the fitted symbolic transition and its recursive rollouts remain in one latent coordinate system. The EMA target encoder is used during Neural JEPA training but is not used to construct the post-hoc symbolic transition dataset. No gradients from the symbolic model are propagated into the frozen encoder.

\paragraph{Stage 3: fitting symbolic dynamics.}
A symbolic vector field is fitted to the frozen latent transitions using the same candidate library and reporting threshold as the jointly learned symbolic model. Its residual transition is
\begin{equation}
\widehat z_{t+1}^{\mathrm{sym}}
=
z_t
+
\Delta t_t
F_{\mathcal E,\alpha}(z_t).
\end{equation}
The symbolic structure $\mathcal E$ and coefficients $\alpha$ are learned by minimising
\begin{equation}
\label{eq:posthoc-symbolic-objective}
\min_{\mathcal E,\alpha}
\quad
\Lpred
+
\lambda_{\mathrm r}\Lroll
+
\lambda_{\mathrm s}\Csym^{\mathrm{train}}.
\end{equation}
The representation regulariser $\RIB$ is absent because the encoder is frozen, and no neural correction is used. Checkpoints are selected using the validation trajectories; among candidates with nearly tied validation risks, the lower-complexity thresholded symbolic law is preferred.

\paragraph{Evaluation.}
One-step latent prediction error is evaluated directly in the frozen Neural JEPA coordinate system. For physical-state rollout evaluation, an affine map from the frozen training latents to $(q,p)$ is fitted using the training trajectories and then applied to test and OOD rollouts. The physical-state variables are used only for this evaluation alignment and are not provided to the symbolic regression procedure.

The post-hoc baseline therefore implements
\begin{equation}
\text{learn predictive coordinates}
\;\longrightarrow\;
\text{freeze the coordinates}
\;\longrightarrow\;
\text{fit symbolic dynamics}.
\end{equation}
By contrast, \ours alternates representation-space and symbolic-dynamics optimisation, allowing the latent coordinates to change so that the induced transition becomes simpler while maintaining predictive adequacy and the prescribed non-collapse constraints.

\subsection{Additional Experimental Results}

\subsubsection{Per-seed equations for Experiment~1}
\label{app:exp1-equations}

After applying the fixed reporting threshold, the selected \ours equations are
\begin{align*}
\text{Seed 7:}\qquad
\dot z_1
&=
-0.806z_2,
&
\dot z_2
&=
0.822z_1+0.177-0.143z_1^2,
\\
\text{Seed 19:}\qquad
\dot z_1
&=
-0.847z_2+0.146-0.085z_1^2,
&
\dot z_2
&=
0.811z_1+0.185-0.143z_1^2+0.116z_1z_2,
\\
\text{Seed 37:}\qquad
\dot z_1
&=
0.867z_2,
&
\dot z_2
&=
-0.835z_1.
\end{align*}

The sign reversal in seed $37$ reflects the non-identifiability of latent orientation rather than a different qualitative mechanism. The cross-coordinate oscillator terms are selected in all three seeds. By contrast, the post-hoc symbolic models select most of the available polynomial and trigonometric library, with mean weighted complexity
\begin{equation*}
26.0\pm3.24.
\end{equation*}

\subsubsection{Representation and alignment metrics}

\begin{table*}[t]
\centering
\scriptsize
\caption{Additional Experiment~1 representation and alignment metrics.}
\label{tab:exp1-additional}
\begin{tabular}{lcccc}
\toprule
Method
&
Train probe $R^2$
&
Test probe $R^2$
&
Effective rank
&
Affine one-step state MSE
\\
\midrule
Neural JEPA
&
$0.849\pm0.067$
&
$0.814\pm0.063$
&
$1.9994\pm0.0004$
&
$0.171\pm0.058$
\\
Post-hoc symbolic
&
$0.849\pm0.067$
&
$0.814\pm0.063$
&
$1.9994\pm0.0004$
&
$0.176\pm0.063$
\\
\ours symbolic
&
$0.396\pm0.295$
&
$0.293\pm0.154$
&
$1.9996\pm0.0001$
&
$0.647\pm0.141$
\\
One-step collapse diagnostic
&
$0.323\pm0.041$
&
$0.308\pm0.027$
&
$1.701\pm0.201$
&
$0.631\pm0.026$
\\
Collapsed fixed-point control
&
$0$
&
$-0.016$
&
$0$
&
$0.931$
\\
\bottomrule
\end{tabular}
\end{table*}

The nonzero affine probe score of the one-step collapse diagnostic does not contradict near-collapse. An affine map can amplify very small state-correlated variations. The decisive collapse diagnostics are the absolute coordinate standard deviation and covariance trace.

\subsubsection{Coefficient and support stability in Experiment~2}

Every permitted symbolic term is selected in all three Experiment~2 seeds. The fitted coefficients have extremely small variation:
\begin{align*}
\text{Incomplete symbolic:}\qquad
\dot q
&=
(0.9607\pm0.0017)p,
&
\dot p
&=
-(0.8382\pm0.0008)\sin(q),
\\
\text{Regularised hybrid:}\qquad
\dot q
&=
(0.9581\pm0.0008)p,
&
\dot p
&=
-(0.7942\pm0.0024)\sin(q),
\\
\text{Unregularised hybrid:}\qquad
\dot q
&=
(0.3324\pm0.0065)p,
&
\dot p
&=
-(0.1718\pm0.0746)\sin(q),
\\
\text{Complete symbolic:}\qquad
\dot q
&=
(0.9830\pm0.0005)p,
&
\dot p
&=
-(0.9553\pm0.0010)\sin(q)
-
(0.3724\pm0.0007)p|p|.
\end{align*}

The complete grammar therefore recovers both the structure and coefficients of the true vector field consistently. Under the incomplete grammar, correction regularisation preserves the symbolic pendulum mechanism; without it, the correction absorbs much of both equations.

\subsubsection{Qualitative and mechanism-specific diagnostics}

\begin{figure*}[t]
\centering
\begin{minipage}[t]{0.49\textwidth}
\centering
\includegraphics[width=\linewidth]{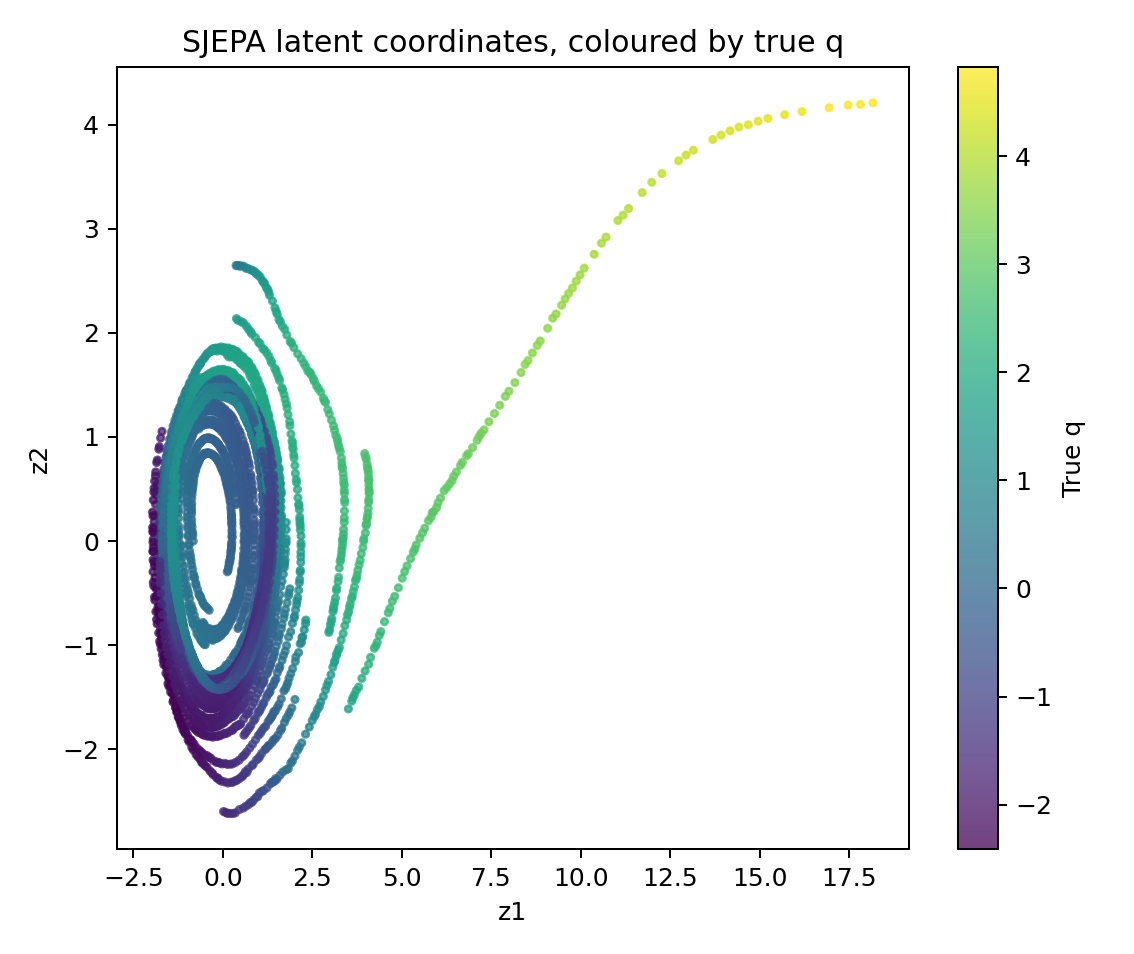}
\vspace{1mm}

\textbf{(a) Example \ours latent geometry}
\end{minipage}
\hfill
\begin{minipage}[t]{0.49\textwidth}
\centering
\includegraphics[width=\linewidth]{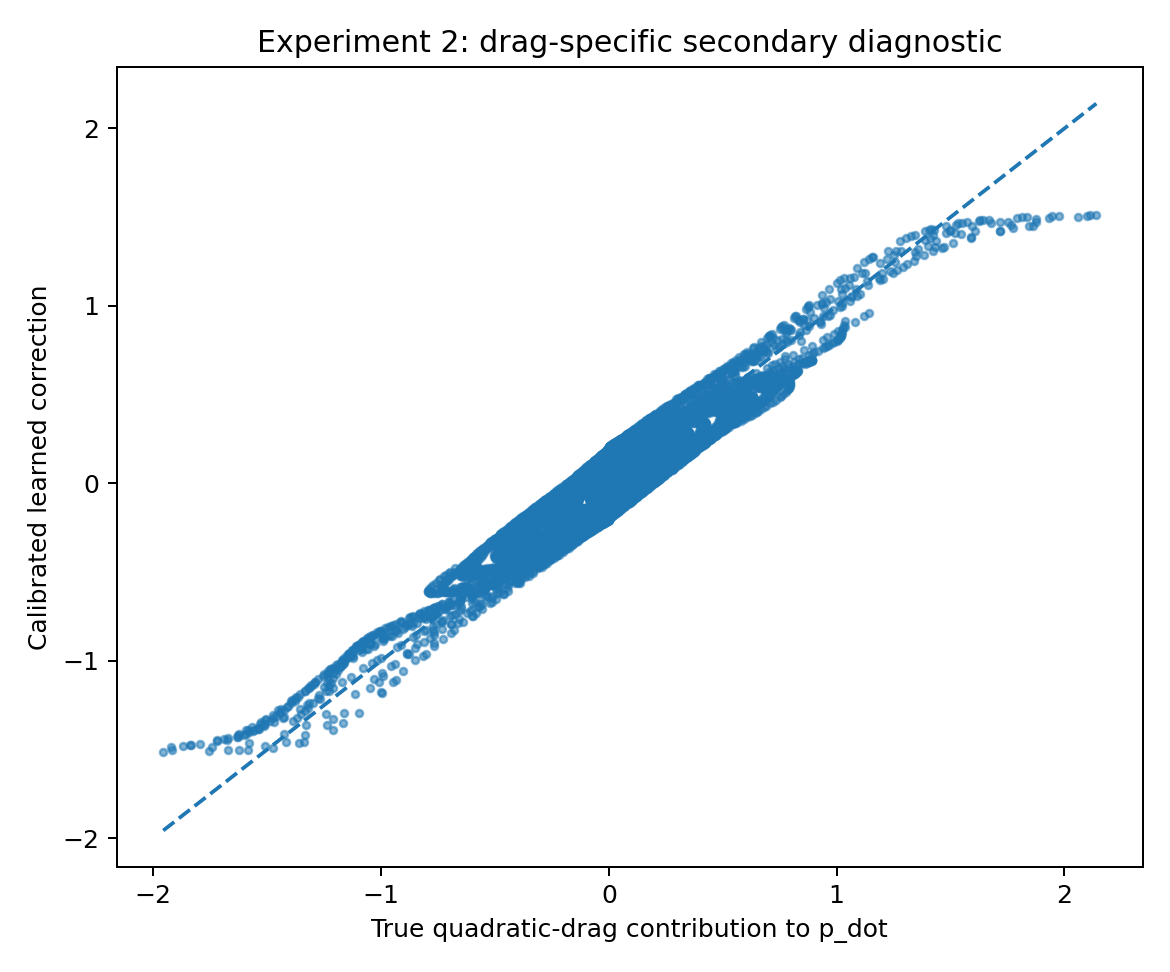}
\vspace{1mm}

\textbf{(b) Drag-specific correction diagnostic}
\end{minipage}
\caption{
Additional qualitative and mechanism-specific diagnostics.
\textbf{(a)} Nonlinear latent geometry in Experiment~1. Each point is an SJEPA embedding $(z_1,z_2)$ coloured by the true pendulum angle $q$. The smooth colour ordering indicates that information about $q$ is retained, while the curved geometry shows that the learned coordinates are not simply affinely aligned with the original state.
\textbf{(b)} Drag-specific diagnostic for Experiment~2 using the regularised hybrid model with the incomplete grammar, whose symbolic component contains $\dot q\propto p$ and $\dot p\propto-\sin(q)$. The panel compares the calibrated $p$-coordinate correction on held-out states with the omitted quadratic-drag term $-0.4p|p|$. By contrast, Figure~\ref{fig:exp2-main}(b) compares the same correction with the complete residual left by the fitted symbolic law, which also includes error arising from imperfect estimation of the $\sin(q)$ coefficient. The dashed line denotes exact recovery.
}
\label{fig:additional-diagnostics}
\end{figure*}

Figure~\ref{fig:additional-diagnostics} collects two secondary diagnostics. Panel~(a) illustrates the geometry of one selected \ours representation in Experiment~1, while panel~(b) evaluates the drag-specific neural correction in Experiment~2. The colour progression follows a curved latent manifold rather than either coordinate axis, suggesting that the true angle $q$ is encoded nonlinearly across both latent dimensions. This visual evidence is qualitative; affine alignment is assessed quantitatively using the probe $R^2$.

The correction is primarily evaluated against the complete residual
\begin{equation}
\label{eq:complete_residual}
r_p
=
f_{\mathrm{true},p}
-
F_{\mathcal E,\alpha,p},
\end{equation}
because this residual contains both the omitted drag and any coefficient error remaining in the fitted symbolic law. Figure~\ref{fig:additional-diagnostics}(b) additionally compares the calibrated correction with the isolated quadratic-drag term. This is therefore a secondary mechanism-specific diagnostic rather than the primary correction target.

\section{Bayesian Inference Details}
\label{app:bayesdetails}

This section provides additional details for the Bayesian formulation in the main text. The encoders remain deterministic throughout: Bayesian uncertainty is introduced only over the latent transition law, its coefficients, transition covariance, and, in the hybrid model, the Gaussian-process correction.

\subsection{Finite-Candidate Symbolic Inference}

Exact inference over all symbolic expression trees is generally intractable. Possible schemes include reversible-jump Markov chain Monte Carlo, sequential Monte Carlo over expression trees, nested sampling, and Bayesian inference over a finite candidate set. A practical implementation for \ours uses symbolic search as a proposal mechanism and subsequently performs Bayesian model comparison over the resulting candidate structures.

Let
\begin{equation*}
\mathcal S_M
=
\left\{
\mathcal E_m
\right\}_{m=1}^{M}
\end{equation*}
denote a finite candidate set generated using candidate-generation data
$\D_{\mathrm{gen}}$. Candidates may be obtained from sparse-library search, genetic programming, beam search, or another symbolic-discovery procedure. Algebraically equivalent or numerically duplicate expressions are canonicalised and removed. To preserve diversity, candidates may be retained from the predictive-error--complexity Pareto set rather than selecting only the lowest-error expressions.

A disjoint evidence set $\D_{\mathrm{evid}}$ is used for Bayesian fitting and model comparison. For each candidate,
\begin{equation}
\label{eq:structure-evidence}
p(\D_{\mathrm{evid}}\mid\mathcal E_m)
=
\int
p(\D_{\mathrm{evid}}\mid\mathcal E_m,\alpha,\Sigma)
p(\alpha\mid\mathcal E_m)
p(\Sigma)
\dd\alpha\,\dd\Sigma.
\end{equation}
The evidence integrates over symbolic coefficients and transition covariance rather than evaluating only one fitted parameter value. It therefore accounts for parameter uncertainty and penalises structures whose apparent fit depends on a narrowly tuned or unnecessarily flexible parameterisation.

The integral in Equation~\eqref{eq:structure-evidence} is generally unavailable in closed form. It may be estimated using posterior sampling, nested sampling, variational inference, or a Laplace approximation. For the latter, let
\begin{equation}
\vartheta_m
=
\left(
\alpha_m,
\lambda_m
\right),
\qquad
\Sigma_m
=
L(\lambda_m)L(\lambda_m)^\top,
\end{equation}
where $\lambda_m$ contains unconstrained parameters defining a lower-triangular Cholesky factor $L(\lambda_m)$ with positive diagonal entries. Thus, $\Sigma_m$ remains positive definite throughout optimisation.

Let
\begin{equation}
\widehat\vartheta_m
=
\argmax_{\vartheta_m}
p(\vartheta_m\mid\D_{\mathrm{evid}},\mathcal E_m)
\end{equation}
be the posterior mode, and let
\begin{equation}
H_m
=
-
\nabla_{\vartheta_m}^{2}
\log
p
\left(
\D_{\mathrm{evid}},
\vartheta_m
\mid
\mathcal E_m
\right)
\bigg|_{\vartheta_m=\widehat\vartheta_m}
\end{equation}
be the negative Hessian of the log joint density at that mode. If
$d_m=\dim(\vartheta_m)$ and $H_m$ is positive definite, the Laplace approximation gives
\begin{align}
\log p(\D_{\mathrm{evid}}\mid\mathcal E_m)
\approx{}&
\log p
\left(
\D_{\mathrm{evid}}
\mid
\mathcal E_m,
\widehat\vartheta_m
\right)
+
\log p
\left(
\widehat\vartheta_m
\mid
\mathcal E_m
\right)
\nonumber\\
&
+
\frac{d_m}{2}\log(2\pi)
-
\frac{1}{2}\log|H_m|.
\label{eq:structure-evidence-laplace}
\end{align}
The log-determinant term accounts for posterior concentration: a structure whose good fit is confined to a narrow parameter region need not receive the same evidence as one that explains the data over a larger plausible region.

Within the selected candidate set, define
\begin{equation}
p(\mathcal E_m\mid\mathcal S_M)
\propto
\exp
\left\{
-\gamma\Csym(\mathcal E_m)
\right\}.
\end{equation}
The posterior structure probabilities are
\begin{equation}
\label{eq:structure-posterior-appendix}
p(\mathcal E_m\mid\D_{\mathrm{evid}},\mathcal S_M)
=
\frac{
p(\D_{\mathrm{evid}}\mid\mathcal E_m)
p(\mathcal E_m\mid\mathcal S_M)
}{
\displaystyle
\sum_{r=1}^{M}
p(\D_{\mathrm{evid}}\mid\mathcal E_r)
p(\mathcal E_r\mid\mathcal S_M)
}.
\end{equation}
For numerical stability, define
\begin{equation}
a_m
=
\log p(\D_{\mathrm{evid}}\mid\mathcal E_m)
-
\gamma\Csym(\mathcal E_m).
\end{equation}
The normalised weight is
\begin{equation}
\label{eq:structure-logsumexp}
w_m
=
\frac{
\exp(a_m-a_{\max})
}{
\displaystyle
\sum_{r=1}^{M}
\exp(a_r-a_{\max})
},
\qquad
a_{\max}
=
\max_{1\le r\le M}a_r.
\end{equation}

For each candidate, the structure-conditional posterior predictive distribution is
\begin{equation}
\label{eq:structure-conditional-predictive}
p
\left(
z_T
\mid
z_C,
\epsinfo,
\D_{\mathrm{evid}},
\mathcal E_m
\right)
=
\int
p
\left(
z_T
\mid
z_C,
\epsinfo,
\mathcal E_m,
\alpha,
\Sigma
\right)
p
\left(
\alpha,
\Sigma
\mid
\D_{\mathrm{evid}},
\mathcal E_m
\right)
\dd\alpha\,\dd\Sigma.
\end{equation}
Bayesian model averaging gives
\begin{equation}
\label{eq:structure-bma}
p
\left(
z_T
\mid
z_C,
\epsinfo,
\D_{\mathrm{evid}},
\mathcal S_M
\right)
=
\sum_{m=1}^{M}
w_m
p
\left(
z_T
\mid
z_C,
\epsinfo,
\D_{\mathrm{evid}},
\mathcal E_m
\right).
\end{equation}

A Monte Carlo approximation first draws
\begin{equation}
J^{(b)}
\sim
\operatorname{Categorical}(w_1,\ldots,w_M),
\end{equation}
then draws
\begin{equation}
\left(
\alpha^{(b)},
\Sigma^{(b)}
\right)
\sim
p
\left(
\alpha,
\Sigma
\mid
\D_{\mathrm{evid}},
\mathcal E_{J^{(b)}}
\right),
\end{equation}
and finally samples or evaluates the corresponding latent transition.

The symbolic-search stage constructs a finite, data-dependent support over which Bayesian model comparison is performed; its original search score is not itself treated as a posterior probability. The resulting weights are approximate posterior probabilities conditional on the selected candidate set $\mathcal S_M$, rather than the exact posterior over the full grammar. Hyperparameters such as $\gamma$, prior scales, and $M$ may be chosen using validation data, while test and OOD data are reserved for final posterior-predictive evaluation. The encoders remain point-estimated and are not included in the Bayesian averaging.

\begin{algorithm}[H]
\caption{Finite-candidate Bayesian symbolic inference}
\label{alg:finite-bayesian-symbolic}
\begin{algorithmic}[1]
\REQUIRE Fixed encoders, candidate-generation data $\D_{\mathrm{gen}}$, evidence data $\D_{\mathrm{evid}}$, symbolic grammar $\mathfrak E$
\STATE Generate candidate structures using $\D_{\mathrm{gen}}$
\STATE Canonicalise expressions and remove duplicate structures
\STATE Retain $M$ candidates from the predictive-error--complexity Pareto set evaluated on $\D_{\mathrm{gen}}$
\FOR{$m=1,\ldots,M$}
    \STATE Infer $p(\alpha,\Sigma\mid\D_{\mathrm{evid}},\mathcal E_m)$
    \STATE Estimate $\log p(\D_{\mathrm{evid}}\mid\mathcal E_m)$
    \STATE Set
    $a_m\leftarrow
    \log p(\D_{\mathrm{evid}}\mid\mathcal E_m)
    -\gamma\Csym(\mathcal E_m)$
\ENDFOR
\STATE Normalise $\{a_m\}_{m=1}^{M}$ using Equation~\eqref{eq:structure-logsumexp}
\STATE Form the posterior predictive distribution using Equation~\eqref{eq:structure-bma}
\RETURN Candidate structures, posterior weights, and posterior predictive model
\end{algorithmic}
\end{algorithm}

\subsection{Gaussian-Process Hybrid Inference}

\paragraph{Output-wise GP correction.}
For a fixed symbolic structure $\mathcal E$, coefficients $\alpha$, and latent output coordinate $j$, define
\begin{equation*}
s_i
=
\left(
z_C^{(i)},
\epsinfo^{(i)}
\right),
\qquad
m_{ij}
=
F_{\mathcal E,\alpha,j}(s_i),
\qquad
r_{ij}
=
z_{T,j}^{(i)}-m_{ij}.
\end{equation*}
Let
\begin{equation}
g_j
\sim
\mathcal{GP}
\left(
0,
k_{\vartheta_{g,j}}
\right),
\end{equation}
and define
\begin{equation}
[K_j]_{ab}
=
k_{\vartheta_{g,j}}(s_a,s_b),
\qquad
C_j
=
K_j+\sigma_j^2I_n.
\end{equation}
The residual vector
\begin{equation*}
\bm r_j
=
(r_{1j},\ldots,r_{nj})^\top
\end{equation*}
has the marginal distribution
\begin{equation}
\label{eq:gp-residual-marginal}
\bm r_j
\mid
\mathcal E,
\alpha,
\vartheta_{g,j},
\sigma_j^2
\sim
\N
\left(
\bm 0,
C_j
\right).
\end{equation}
Consequently, its negative log marginal likelihood is
\begin{equation}
\label{eq:gp-nlml}
-\log
p
\left(
\bm r_j
\mid
\mathcal E,
\alpha,
\vartheta_{g,j},
\sigma_j^2
\right)
=
\frac{1}{2}
\bm r_j^\top
C_j^{-1}
\bm r_j
+
\frac{1}{2}
\log|C_j|
+
\frac{n}{2}\log(2\pi).
\end{equation}
The quadratic term measures how well the GP explains systematic residual structure left by the symbolic law. The log-determinant term controls the covariance flexibility used to explain the data. Their balance concerns residual fit and GP complexity, but does not by itself identify a unique symbolic--GP allocation.

\paragraph{Joint symbolic--GP posterior.}
Let $g=(g_1,\ldots,g_{d_z})$ and let $\vartheta_g$ collect the GP hyperparameters. The joint posterior has the schematic form
\begin{equation}
\label{eq:bayesian-hybrid-posterior}
p(\mathcal E,\alpha,g,\vartheta_g,\Sigma\mid\D_Z)
\propto
p(\D_Z\mid\mathcal E,\alpha,g,\Sigma)
p(g\mid\vartheta_g)
p(\vartheta_g)
p(\alpha\mid\mathcal E)
p(\Sigma)
p(\mathcal E).
\end{equation}
Uncertainty in $\mathcal E$ concerns symbolic structure; uncertainty in $\alpha$ concerns its numerical coefficients; uncertainty in $g$ concerns systematic residual dynamics; and $\Sigma$ represents transition variability remaining after conditioning on the symbolic law and correction. The symbolic--GP allocation remains dependent on the structure prior, coefficient priors, GP kernel and amplitude priors, and transition-noise prior.

Integrating out $g$ gives the GP marginal likelihood in Equation~\eqref{eq:gp-nlml}. One may then infer or average over $\mathcal E$, $\alpha$, kernel hyperparameters, and $\Sigma$ using posterior sampling, variational approximations, Laplace approximations, or a finite candidate ensemble.

\paragraph{Posterior correction at a new input.}
For
\begin{equation*}
s_\star
=
(z_C,\epsinfo),
\end{equation*}
define
\begin{equation*}
\bm k_{\star,j}
=
\left(
k_{\vartheta_{g,j}}(s_\star,s_1),
\ldots,
k_{\vartheta_{g,j}}(s_\star,s_n)
\right)^\top,
\qquad
k_{\star\star,j}
=
k_{\vartheta_{g,j}}(s_\star,s_\star).
\end{equation*}
Conditional on fixed symbolic and GP parameters,
\begin{equation}
\label{eq:gp-correction-posterior}
g_j(s_\star)
\mid
\D_Z,
\mathcal E,
\alpha,
\vartheta_{g,j},
\sigma_j^2
\sim
\N
\left(
\mu_{g_j}(s_\star),
v_{g_j}(s_\star)
\right),
\end{equation}
where
\begin{align}
\mu_{g_j}(s_\star)
&=
\bm k_{\star,j}^\top
C_j^{-1}
\bm r_j,
\\
v_{g_j}(s_\star)
&=
k_{\star\star,j}
-
\bm k_{\star,j}^\top
C_j^{-1}
\bm k_{\star,j}.
\end{align}
The posterior mean supplies the data-supported correction to the symbolic law, while the posterior variance typically becomes larger in regions weakly supported by the latent transition data, subject to the selected kernel and hyperparameters.

For vector-valued target embeddings, one may use independent output-wise GPs, a shared kernel with output-specific parameters, or a matrix-valued kernel that models dependence between latent coordinates. Independent output-wise models are computationally simplest, whereas multi-output kernels can represent correlated transition uncertainty.

\subsection{Posterior Predictive Rollouts}

Posterior predictive rollouts propagate uncertainty recursively through the learned latent dynamics. For rollout sample $b$, first draw one coherent transition model:
\begin{equation}
\label{eq:rollout-model-draw}
\left(
\mathcal E^{(b)},
\alpha^{(b)},
g^{(b)},
\vartheta_g^{(b)},
\Sigma^{(b)}
\right)
\sim
p
\left(
\mathcal E,
\alpha,
g,
\vartheta_g,
\Sigma
\mid
\D_Z
\right).
\end{equation}
Starting from $z_t^{(b)}=z_t$, recursively sample
\begin{equation}
\label{eq:bayesian-rollout-step}
z_{t+h+1}^{(b)}
\sim
\N
\left(
F_{\mathcal E^{(b)},\alpha^{(b)}}
\left(
z_{t+h}^{(b)},
\epsinfo_{t+h}
\right)
+
g^{(b)}
\left(
z_{t+h}^{(b)},
\epsinfo_{t+h}
\right),
\Sigma^{(b)}
\right),
\end{equation}
for $h=0,\ldots,K-1$.

Equation~\eqref{eq:bayesian-rollout-step} gives the direct-transition formulation. For a vector-field realisation, its mean is replaced by the corresponding integrated transition, for example
\begin{equation*}
z_{t+h}^{(b)}
+
\Delta t_{t+h}
\left[
F_{\mathcal E^{(b)},\alpha^{(b)}}
\left(
z_{t+h}^{(b)},
\epsinfo_{t+h}
\right)
+
g^{(b)}
\left(
z_{t+h}^{(b)},
\epsinfo_{t+h}
\right)
\right]
\end{equation*}
under forward Euler.

The same sampled structure, coefficients, and GP function are retained throughout a rollout. This preserves the interpretation of epistemic uncertainty as uncertainty about one underlying transition model. Resampling the symbolic law or GP independently at every step would instead introduce artificial temporal variation in the model itself. Transition noise may still be sampled separately at each step because it represents aleatoric variability conditional on the sampled dynamics.

Repeating Equations~\eqref{eq:rollout-model-draw}--\eqref{eq:bayesian-rollout-step} produces an empirical approximation to
\begin{equation}
p
\left(
z_{t+1:t+K}
\mid
z_t,
\epsinfo_{t:t+K-1},
\D_Z
\right).
\end{equation}
The resulting distribution reflects uncertainty over symbolic structure, coefficients, systematic correction, transition covariance, and the uncertain states visited during recursion.

Calibration should be evaluated separately for one-step and multi-step prediction. Even a well-calibrated one-step model can become miscalibrated over long horizons because uncertain states are repeatedly fed back into the transition, model misspecification compounds, and trajectories may enter regions poorly represented in the training data. Appropriate diagnostics include horizon-wise log predictive density, empirical interval coverage, calibration error, and task-level decision quality under posterior-predictive planning.

\end{document}